\pdfoutput=1

\documentclass[11pt]{article}
 
\usepackage[final]{acl}

\usepackage{times}
\usepackage{latexsym}
\usepackage{tcolorbox}

\usepackage[T1]{fontenc}

\usepackage[utf8]{inputenc}

\usepackage{microtype}

\usepackage{inconsolata}

\usepackage{graphicx}

\usepackage{marvosym}

\usepackage{caption}
\usepackage{subcaption}
\usepackage{algorithm}
\usepackage{algpseudocode}
\usepackage{amsmath}
\usepackage{mathtools}

\usepackage{comment}
\usepackage[]{hyperref}
\usepackage{subcaption}
\usepackage{textcomp}
\usepackage{xspace}
\usepackage{enumitem}
\usepackage[normalem]{ulem}
\usepackage{makecell} 
\usepackage{cleveref}

\renewcommand{\emph}[1]{\textit{#1}}

\NewDocumentCommand{\smallpm}{m o}{%
  #1
  \IfValueT{#2}{
    ${}{\scriptstyle\pm#2}$
  }%
}

\title{Slim-SC: Thought Pruning for Efficient Scaling with Self-Consistency}

\author{
Colin Hong\textsuperscript{1,*}, 
Xu Guo\textsuperscript{2,*\Letter}, 
Anand Chaanan Singh\textsuperscript{1}, \\
\textbf{Esha Choukse\textsuperscript{3}}, 
\textbf{Dmitrii Ustiugov\textsuperscript{1}} \\
\textsuperscript{1}NTU Singapore \quad
\textsuperscript{2}KTH Royal Institute of Technology \quad
\textsuperscript{3}Microsoft \\
\texttt{\{hong0259,xu008\}@e.ntu.edu.sg}, 
\texttt{dmitrii.ustiugov@ntu.edu.sg} \\
\textsuperscript{*} Equal contribution. \quad
\textsuperscript{\Letter} Corresponding author.
}

\ifdefined\RELEASE
  \newcommand{\ignore}[1]{}
  \newcommand{\fixme}[1]{}
  \newcommand{\xu}[1]{}
  \newcommand{\col}[1]{}
  \newcommand{\dmi}[1]{}
  \newcommand{\cha}[1]{}
  \newcommand{\esha}[1]{}

  \newcommand{\TODO}[1]{}

\else
  \newcommand{\ignore}[1]{}
  \newcommand{\fixme}[1]{{\textcolor{red}{[~FIXME:~#1~]}}}
  \newcommand{\xu}[1]{{\textcolor{brown}{[~X:~#1~]}}}
  \newcommand{\col}[1]{{\textcolor{purple}{[~Co:~#1~]}}}
  \newcommand{\dmi}[1]{{\textcolor{blue}{[~D:~#1~]}}}
  \newcommand{\cha}[1]{{\textcolor{orange}{[~CH:~#1~]}}}
  \newcommand{\esha}[1]{{\textcolor{teal}{[~ES:~#1~]}}}
  \newcommand{\TODO}[1]{{\textcolor{red}{TODO:~#1}}}
\fi

\begin{document}
\maketitle
\begin{abstract}
Recently, Test-Time Scaling (TTS) has gained increasing attention for improving LLM reasoning performance at test time without retraining the model. A notable TTS technique is Self-Consistency (SC), which generates multiple reasoning chains in parallel and selects the final answer via majority voting. While effective, the order-of-magnitude computational overhead limits its broad deployment. Prior attempts to accelerate SC mainly rely on model-based confidence scores or heuristics with limited empirical support. 
For the first time, we theoretically and empirically analyze the inefficiencies of SC and reveal actionable opportunities for improvement. Building on these insights, we propose Slim-SC, a step-wise pruning strategy that identifies and removes redundant chains using inter-chain similarity at the thought level.
Experiments on three STEM reasoning datasets and two recent LLM architectures show that Slim-SC reduces inference latency and KVC usage by up to 45\% and 26\%, respectively, with R1-Distill, while maintaining or improving accuracy, thus offering a simple yet efficient TTS alternative for SC.

\end{abstract}

\section{Introduction}
In recent years, advances in Large Language Models (LLMs) have mainly come from training-time scaling, through more parameters, bigger datasets, and more compute~\cite{gpt3,PaLM,gpt4,llama,deepseekr1}. Such scale has unlocked emergent reasoning abilities, e.g., through Chain-of-Thought (CoT) prompting~\cite{CoT}, which prompts the LLM to generate intermediate thinking steps to enhance answer accuracy. Building on this foundation, a new scaling paradigm known as Test-Time Scaling (TTS)~\cite{SC-CoT,optimal-scaling,tts-resampling,s1} has gained traction, which aims to further enhance reasoning performance by allocating additional computation at inference time,  without modifying the model’s parameters.

\begin{figure*}[t!]
\begin{subfigure}[b]{0.32\textwidth}
  \includegraphics[width=\columnwidth]{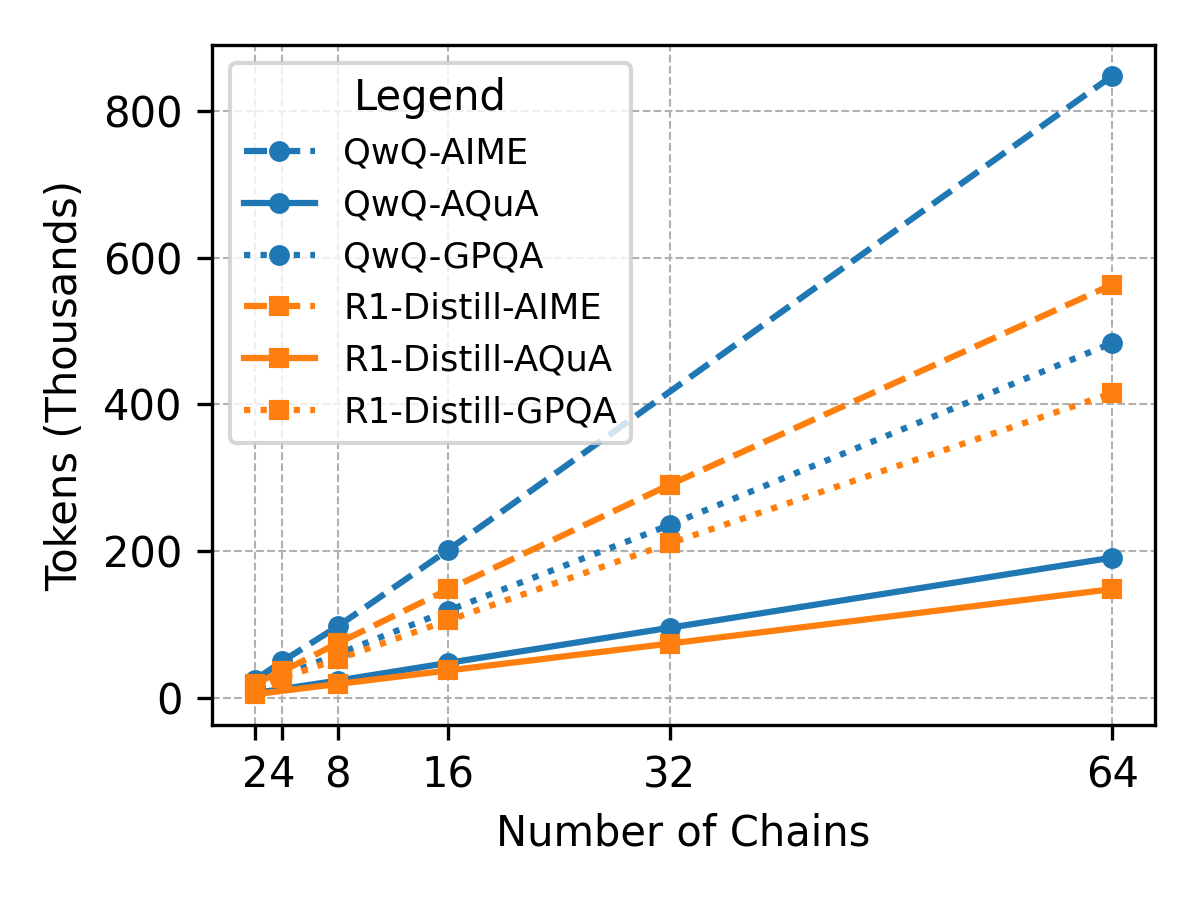}
  \caption{Computational cost scaling.
  }
  \label{fig:token-cost}
\end{subfigure}
\hfill
\begin{subfigure}[b]{0.33\textwidth}
\includegraphics[width=\columnwidth]{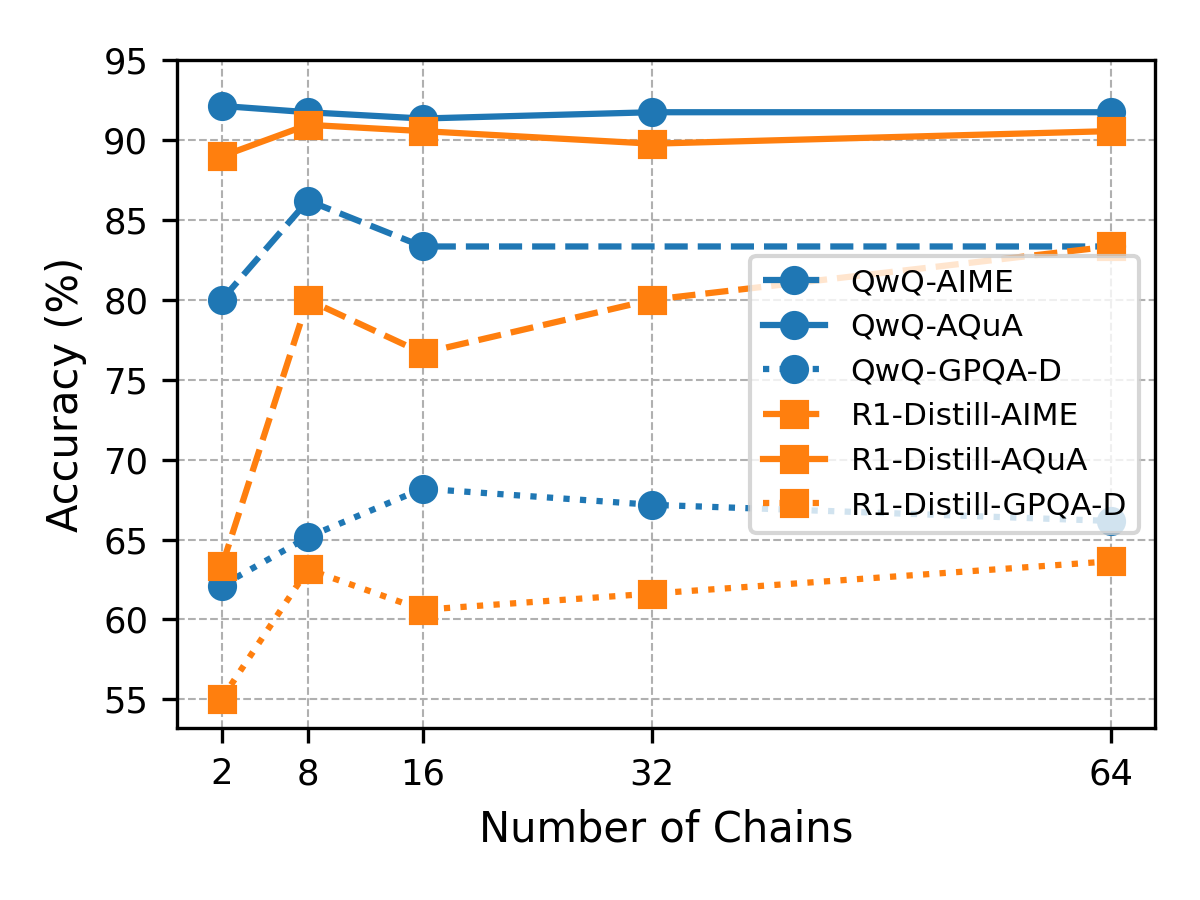}
\caption{Accuracy across different datasets. 
}
\label{fig:motivation-accuracy}
\end{subfigure}
\hfill
\begin{subfigure}[b]{0.32\textwidth}
\includegraphics[width=\columnwidth]{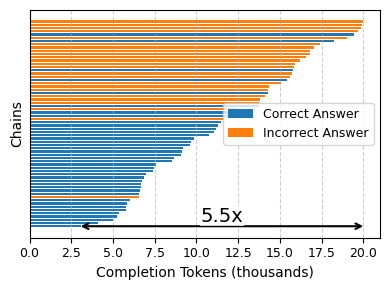}
\caption{SC ($N=64$) for R1-Distill on \texttt{AIME}.
}
\label{fig:completion-diff}

\label{fig:completion-diff}
\end{subfigure}
\caption{\textbf{Problem with Self-Consistency.} (a) and (b): Diminishing returns from scaling test-time computations. Accuracy generally improves with more chains, but plateaus and diminishes in gains, especially for easier datasets such as \texttt{AQuA-RAT}. (c) SC waits for all chains, including longer incorrect ones. (Appendix~\ref{appendix:completion-tokens-cdf} shows the distribution of completion tokens for correct vs. incorrect chains on the \texttt{AIME} dataset. Appendix \ref{appendix:wait-for-all} gives more examples.)
}
\label{fig:sc-problems}
\end{figure*}

Current TTS methods can generally be divided into two categories. Sequential scaling, such as Least-to-Most prompting~\cite{leasttomost}, directs later thinking steps based on earlier reasoning results~\cite{self-refine,khot2023decomp}. While this can improve local coherence and reasoning depth, it is also vulnerable to error propagation. Parallel scaling, such as Best-of-N (BoN)~\cite{BoN} and Self-Consistency (SC)~\cite{SC-CoT}, takes a different route. They sample many reasoning chains independently and aggregate the final answer through a fusion mechanism. 
These methods aim to increase diversity among reasoning paths to improve the chance of reaching a correct answer. However, the extra samples raise non-trivial latency and cost (Fig.~\ref{fig:token-cost}).

Recent attempts to speed up parallel scaling mainly focus on sampling fewer chains~\cite{ASC,ESC, DSC,RASC}: they generate chains in batches, perform an early vote, and stop if the batch already agrees. This strategy shifts part of the computation into a sequential loop, where each new batch must wait for the previous vote, thereby cutting total compute but often increasing end-to-end latency. Moreover, our study shows that correct and incorrect chains often exhibit different patterns (Fig.~\ref{fig:completion-diff}, Fig.~\ref{fig:similar-thoughts-lead-to-same-answer}). Early agreement inside any single batch, therefore, does not guarantee accuracy improvement for SC. Results in Tab.~\ref{tab:accuracy-comparison} confirm this and are consistent with ~\citet{ESC}.

Some approaches maintain a fixed number of reasoning chains while selectively pruning each chain as it unfolds.
They typically rely on a pretrained reward model to estimate confidence at intermediate steps~\cite{fast-BoN,ST-BoN}. Others adopt information-theoretic metrics, including entropy trajectory~\cite{entropy}, which monitors reasoning utility, and inter-chain distances~\cite{RPC} to detect outlier chains. However, reward-based pruning heavily depends on the accuracy of the confidence estimation model, which may not always be available. Meanwhile, pruning outliers based on inter-chain distances can be unreliable, as our study shows that correct and incorrect chains each form distinct clusters (Fig.~\ref{fig:similar-thoughts-lead-to-same-answer}).

To improve the efficiency of Self-Consistency (SC) without compromising accuracy, we propose \textbf{Slim-SC}, a step-wise pruning method that removes reasoning chains based on inter-chain similarity at the thought level.
Our motivation study shows that \textbf{(1)} scaling SC chains indeed increases the presence of accurate answers in the candidate pool, but they are often overwhelmed by a larger number of incorrect ones, leading to the wrong final answer and, hence, lower accuracy (Fig.~\ref{fig:oracle-actual-diff}); \textbf{(2)} chains that lead to incorrect answers tend to be longer \cite{hassid2025dontoverthinkitpreferring}, as compared to the ones leading to correct answers, hence increasing the latency of SC.
Since chains with highly similar intermediate thoughts almost always converge to the same final answer, pruning redundant chains before voting yields a triple benefit: lowering compute cost, latency, and improving accuracy when incorrect chains are abundant. 
Because correct and incorrect chains are well-separated, our similarity-based pruning primarily eliminates redundancy within each cluster, thereby preserving overall accuracy.
Driven by these insights, Slim-SC employs two mechanisms: first, it introduces a warm-up delay, ensuring meaningful pruning, since reasoning chains typically start by problem restatement; second, it applies a similarity-based pruning rule at each thought to decide which chain to retain when two chains become highly similar. We experimentally evaluate both random selection and diversity-based selection for this pruning rule. 
Extensive experiments on three STEM reasoning datasets and two recent LLM architectures confirm that Slim-SC maintains or improves the accuracy of SC and outperforms existing pruning methods. 
More importantly, Slim-SC substantially reduces the latency of SC (up to 45\%), as well as the generated token number (up to 34\%), and KVC usage (up to 26\%). 
Slim-SC is thus a simple yet effective and broadly compatible approach, suitable for integration into various parallel scaling frameworks using SC or similar methods \cite{ESC,BoN,fast-BoN}.

Our code is available at \url{https://github.com/hyscale-lab/slimsc}.

\section{Motivation}

\subsection{Self-Consistency}\label{appendix:self-consistency}

Given an input $x$ with groundtruth $y$, an LLM $\theta$ under CoT prompting generates a reasoning chain $c$ for $x$. SC enables test-time scaling by sampling a set of $N$ independent reasoning paths $\mathcal{R}_N=\{c_i\}_{i=1}^{N}\stackrel{\text{i.i.d.}}{\sim}P_\theta(c\mid x)$ and maps each reasoning chain to an answer $\hat{y}_i=g(c_i)$. The final prediction $\tilde{y}$ is obtained by \emph{plurality voting} (i.e., selecting the most frequent answer):
\[
\tilde{y} \;=\;\operatorname*{arg\,max}_{y}\sum_{i=1}^{N}\mathbf 1\!\bigl[\hat{y}_i=y\bigr].
\]

\noindent SC improves accuracy only if correct chains occur more often than incorrect chains in the candidate pool. Let $p_y$ be the probability of generating the correct answer $y$, and $\{p_{y'_j}\}$ be the probabilities for each incorrect answer $y'_j$. SC is expected to improve accuracy if the correct reasoning path has a probabilistic advantage, i.e., $p_y > \max_j(p_{y'_j})$. Even when $p_y < 0.5$, as long as $p_y$ exceeds every $p_{y'_j}$, increasing $N$ will raise the probability that $y$ wins the plurality vote.

\subsection{Motivation Study} \label{motivation-study}

To characterize the costs and benefits of standard SC, we investigate its performance across various datasets and models. We aim to quantify how accuracy and computational cost scale with the number of SC chains ($N$). Furthermore, we analyze the nature of these chains, particularly focusing on their completion lengths and semantic similarity, to identify sources of inefficiency and opportunities for optimization. Specifically, we examine the similarity patterns within and between chains that lead to correct versus incorrect final answers,  referred to as \emph{correct} and \emph{incorrect} chains respectively.
For this characterization, we primarily use \texttt{AIME-2024}, \texttt{GPQA Diamond}, and \texttt{AQuA-RAT} datasets. The LLMs evaluated are Qwen-QwQ-32B and DeepSeek-R1-Distill-Qwen-14B, which we will shorthand as QwQ and R1-Distill for the rest of the paper. 

\begin{figure*}[t!]
\begin{subfigure}[b]{0.49\textwidth}
\includegraphics[width=\columnwidth]{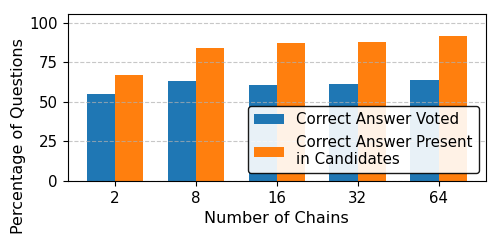}
\caption{R1-Distill on \texttt{GPQA Diamond}: $\%$ of correct answers appearing in the candidate set vs. being selected as the final answer.}
\label{fig:oracle-actual-diff}
\end{subfigure}
\hfill
\begin{subfigure}[b]{0.49\textwidth}
\includegraphics[width=\columnwidth]{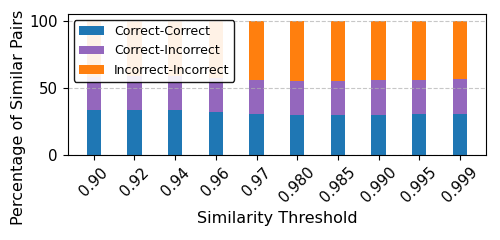}
\caption{Pairwise similarity of chains using R1-Distill on \texttt{GPQA Diamond}.}
\label{fig:similar-thoughts-lead-to-same-answer}
\end{subfigure}
\caption{\textbf{Opportunities.} (a) Many correct chains go unselected, suggesting substantial room for accuracy improvement. (b) Correct and incorrect chains form separate clusters, with high internal similarity within cluster.
}
\label{fig:motivation-insights}
\end{figure*}
\subsection{Problems with Scaling SC Chains}\label{problems-scaling-sc}
\noindent\textbf{Diminishing returns and massive resource waste in scaling SC chains.}
SC can substantially improve reasoning accuracy over single-chain CoT. However, this enhanced performance is accompanied by a steep computational cost that accompanies a more gradual accuracy gain as the number of chains increases.
As observed in Fig. \ref{fig:token-cost}, the mean total completion tokens generated per question scales linearly with the number of chains ($N$). For instance, using R1-Distill on the \texttt{AIME} dataset, increasing $N$ from 2 to 64 results in a more than 30-fold increase in token generation. While accuracy generally improves with more chains (Fig. \ref{fig:motivation-accuracy}), the gains tend to plateau, especially for easier datasets like \texttt{AQuA-RAT}, making the continued increase in computational cost less justifiable.

\noindent\textbf{Short correct chains have to wait for longer incorrect chains.}
Standard SC requires generating $N$ complete reasoning paths before majority voting can occur. This \emph{wait-for-all} strategy introduces significant inefficiencies \cite{hassid2025dontoverthinkitpreferring}. Fig. \ref{fig:completion-diff} exemplifies this with R1-Distill on an \texttt{AIME} Q24 problem ($N=64$): a correct answer might be derived from a short chain, yet the system must wait for all chains, including potentially much longer chains that yield incorrect answers (e.g., a chain can be six times longer than the shortest correct one). Such delays for unhelpful chains contribute to substantial time and computational overhead.

\subsection{Opportunities for Improvement}
\label{sec:opp-for-improvement}
\noindent\textbf{Correct chains can be outvoted by incorrect ones.}
Even when a correct reasoning path is generated, majority voting can miss it. Fig. \ref{fig:oracle-actual-diff} (R1-Distill on \texttt{GPQA Diamond}) shows a persistent gap between the \emph{Correct Answer Present in Final Candidates} (Ideal Accuracy) and the \emph{Correct Answer Voted} (Actual Accuracy). For $N=64$, while the correct answer is present in candidates 91.9\% of the time, it is only voted as the final answer 63.6\% of the time. This \textit{opportunity gap} suggests that \textit{the presence of noisy or unhelpful chains can mislead the voting process}. 
Thus, merely increasing the number of chains is insufficient; improving the set of effective candidate chains for voting is crucial.

\noindent\textbf{Correct and incorrect chains are semantically distant}.
Fig. \ref{fig:similar-thoughts-lead-to-same-answer} presents an analysis of the pairwise similarity of chains for R1-Distill on \texttt{GPQA Diamond} with $N=64$ (computation details in Appendix \ref{appendix:prop_of_sim_chain}, and more examples in Fig.~\ref{fig:pairwise-similarities}).
Key observations:
\begin{itemize}[leftmargin=0cm,itemindent=.4cm,labelwidth=\itemindent,labelsep=0cm,align=left]
\setlength\itemsep{0pt}
    \item \textbf{High Intra-Category Similarity:} Both correct and incorrect chains demonstrate strong internal similarity. For instance, at the 0.98 threshold, most highly similar pairs are either `Correct–Correct' (30\%) or `Incorrect–Incorrect' (45\%), suggesting that reasoning chains leading to the same outcome tend to follow comparable trajectories.
    \item \textbf{Low Cross-Category Similarity:} 
    Correct and incorrect chains rarely overlap semantically. Only 25\% of highly similar pairs are `Correct–Incorrect', indicating that the two groups form largely distinct clusters in the representation space. As shown in Fig.~\ref{fig:pairwise-similarities}, this proportion is even smaller for both R1-Distill and QwQ models on the AQuA dataset.

    \item \textbf{Increasing Threshold Amplifies the Clustering of Incorrect Chains:} Raising the similarity threshold increases the proportion of similar `Incorrect-Incorrect' pairs. For instance, at the 0.98 threshold, 45\% of highly similar pairs are `Incorrect-Incorrect', compared to only 30\% that are `Correct-Correct'. As illustrated in Fig.~\ref{fig:pairwise-similarities}, this effect is particularly pronounced for R1-Distill relative to QwQ. These results suggest that flawed reasoning converges more consistently on similar (wrong) logic than valid reasoning does on the same correct logic.
    
\end{itemize}
These patterns create an opportunity to prune redundancy without harming diversity. The key insight is that correct and incorrect chains are semantically distant, so a high-similarity pruning strategy will primarily remove chains \emph{within} a category rather than across categories. This means pruning is unlikely to mistakenly remove a correct chain that happens to be similar to an incorrect one. Instead, the primary effect of pruning is removing \emph{redundancy} by eliminating chains that are very similar to others, aiming to preserve a smaller, more diverse set of unique reasoning paths for the final vote.

This opportunity directly targets the multiple inefficiencies of standard SC. The high resource cost and diminishing returns of scaling $N$ (Fig.~\ref{fig:sc-problems}a,b), combined with the severe latency bottleneck of the \emph{wait-for-all} approach (Fig.~\ref{fig:sc-problems}c), underscore the need for a smarter strategy than simply generating more chains. By managing redundancy through semantic pruning, we can create a method that simultaneously reduces computational waste and the \emph{wait-for-all} problem, and potentially improves the quality of the final vote by preserving diversity.

\section{The Thought-Pruning Method}
Fig. \ref{fig:similar-thoughts-lead-to-same-answer} reveals that semantically similar intermediate thought pairs converge on identical outcomes: similar correct chains lead to the same correct answer, while similar incorrect chains lead to incorrect results. This redundancy suggests an opportunity for efficiency: by pruning chains that exhibit highly similar thinking processes to others, we can potentially reduce computational overhead while preserving, or even enhancing, the diversity of unique reasoning paths crucial for accurate decision-making. Our goal is to maintain a compact set of diverse chains, pruning away those that offer little new information, particularly if those similar chains might collectively reinforce an incorrect answer due to a lack of diverse perspectives.

\subsection{The Pruning Delay Mechanism} \label{thought_pruning_delay}
A key consideration in our pruning strategy is the warm-up phase of reasoning. 
CoT reasoning often begins with common preliminary thinking segments, such as restating the problem, defining variables, or outlining a general approach. These initial segments can appear highly similar across chains even if their subsequent core reasoning strategies diverge significantly. Consequently, applying similarity-based pruning too early might prematurely eliminate chains that would later develop unique and valuable reasoning paths. To mitigate this, we introduce a \textit{pruning delay}: similarity checks and pruning actions are deferred until chains have progressed beyond a set number of generation steps\footnote{In our experiments, we define each such step as an interval of 3s. In each step, the system attempts to generate and process thought segments (as defined by stop-word delimiters in Section 4.2) from every currently active reasoning chain.}(e.g., \texttt{num\_steps\_to\_delay\_pruning} analysis intervals, such as 20, as implemented in our system).
This delay allows chains to sufficiently develop their distinct approaches before being evaluated for similarity. 

\subsection{The Pruning Strategy and Algorithm}

Once the pruning delay period has passed, our pruning mechanism operates as follows: at each subsequent analysis interval, newly generated \emph{thoughts} (split by stop words like 'Alternatively', 'Wait', 'Another', etc.) are extracted from all active chains. Each new thought from a chain $c_i$ is embedded into a vector representation $e_i$. This embedding $e_i$ is then compared against the embeddings of previously generated thoughts from all \emph{other} active chains $c_j$ ($j \neq i$) stored in an efficient search index (FAISS) \cite{faiss} using cosine similarity. If the maximum similarity score between $e_i$ and any thought $e_j$ from a different chain $c_j$ exceeds a predefined threshold $\tau_{sim}$, i.e., $sim(e_i, e_j) > \tau_{sim}$, then chains $c_i$ and $c_j$ are considered a \emph{similar pair.} One of these chains is then selected for pruning. Based on Fig. \ref{fig:similar-thoughts-lead-to-same-answer}, which shows that highly similar thoughts (e.g., similarity > 0.9) frequently lead to the same final answer, we select $\tau_{sim} > 0.9$. This high threshold ensures we primarily prune chains that are semantically very close, minimizing the risk of losing valuable, distinct reasoning paths.

Let $C = \{c_1, c_2, \dots, c_k\}$ be the set of $k$ active reasoning chains. If a new thought from $c_i$ is deemed highly similar to an existing thought in $c_j$, we employ one of the following strategies to select which chain to prune:

\begin{itemize}[leftmargin=0cm,itemindent=.4cm,labelwidth=\itemindent,labelsep=0cm,align=left]
\setlength\itemsep{0pt}
    \item \textbf{Random Pruning ($\mathcal{P}_{rand}$):} Upon identifying a similar pair $(c_i, c_j)$, one chain is chosen uniformly at random to be pruned. This strategy is simple and provides a baseline for comparison. Formally, we prune $c_x$ where $x$ is selected randomly from $\{i, j\}$.
    \item \textbf{Diversity-based Pruning ($\mathcal{P}_{div}$):} This strategy aims to retain the chain that exhibits greater \textit{internal diversity} among its own thoughts. For each chain $c_k$ in the similar pair, we calculate its internal diversity as the mean pairwise cosine similarity of all thought embeddings within that chain, denoted as $S_{internal}(c_k)$.
    \begin{equation}
        S_{internal}(c_k) =\frac{1}{\binom{m_k}{2}} \sum_{1 \le l < n \le m_k} \text{sim}(e_{k,l}, e_{k,n})
        \label{eq:internal_similarity}
    \end{equation}
    where $\binom{m_k}{2}$ is the number of distinct pairs of thoughts, assuming $m_k \ge 2$. A lower $S_{internal}(c_k)$ indicates greater internal diversity (i.e., thoughts within the chain are less similar to each other).
    If $S_{internal}(c_i) > S_{internal}(c_j)$, $c_i$ is pruned (as it is less internally diverse).
\end{itemize}
In all cases, a chain is only pruned if at least one other active chain remains, ensuring the process does not terminate prematurely.

The general procedure for Slim-SC with thought pruning is described in Algorithm \ref{alg:slim_sc_pruning}, with the chain selection logic detailed in Algorithm \ref{alg:select_prune_chain}.

\section{Experimental Setup}

\subsection{Datasets and Models}

We evaluate on three benchmarks: \texttt{GPQA Diamond} (graduate-level STEM questions requiring complex reasoning) \cite{gpqa}, \texttt{AIME-2024} (high-difficulty math problems from a national competition) \cite{aime}, and \texttt{AQuA-RAT} (algebraic word problems testing multi-step numerical reasoning) \cite{aqua}.
The models used in our experiments are DeepSeek-R1-Distill-Qwen-14B (R1-Distill)~\cite{deepseek} and Qwen-QwQ-32B (QwQ)~\cite{qwq}. For standard Self-Consistency (SC), we first search for the optimal number of chains ($N$) for each dataset and model. The specific values of $N$ used for each configuration are detailed in Table~\ref{tab:optimal-n}. All pruning baselines and our proposed Slim-SC are evaluated in these optimal $N$ settings.
To measure semantic similarity between thought segments, we first embed all textual outputs using the \texttt{all-mpnet-base-v2}\footnote{\url{https://huggingface.co/sentence-transformers/all-mpnet-base-v2}} Sentence Transformers model \cite{sbert}. 
Subsequently, we compute the cosine similarity between the resulting vector embeddings to obtain pairwise similarity scores.

The embedding and FAISS-based similarity search operations were handled on a separate GPU as model inference to ensure proper isolation, but can be run on the same GPU as model inference. We estimate the computational overhead of these operations to be minimal. The embedding model occupies less than 2\% of the GPU memory, and the end-to-end latency from embedding and FAISS search adds an estimated upper-bound delay of only 0.3\%. This negligible cost is heavily outweighed by the substantial latency reductions gained from pruning.

\subsection{Baseline Methods}
We compare Slim-SC against the following established baselines:

\begin{enumerate}[leftmargin=0cm,itemindent=.4cm,labelwidth=\itemindent,labelsep=0cm,align=left]
\setlength\itemsep{0pt}
    \item \textbf{Chain-of-Thought (CoT)} \cite{CoT}
    : This baseline 
    uses standard CoT prompting with greedy decoding, generating a single reasoning path to produce the final answer.
    \item \textbf{Self-Consistency (SC)} \cite{SC-CoT}
    : The standard Self-Consistency sampling approach. SC generates $N$ independent reasoning paths and uses majority voting, which is our primary accuracy baseline.
    \item \textbf{Early-Stopping Self-Consistency (ESC)} \cite{ESC} 
    sequentially generates chains in batches (windows) of $W$ chains until it reaches a consensus or until it generates the maximum $N$ chains (followed by majority voting).
    For a fair comparison, we set $N$ equal to the optimal values of SC and then search for optimal $W$ for ESC. More details are in the Appendix~\ref{appendix:ESC-reproduction}.
    \item \textbf{Concise CoT with Self-Consistency (CCoT-SC)} \cite{concise-cot}
    is a na\"ive baseline adding the instruction "Be concise." to the prompt in the standard SC
    to shorten reasoning paths. It serves as a straightforward prompting baseline to improve SC efficiency, which mainly depends on the LLMs' instruction-following capability.
\end{enumerate}

\section{Results and Discussion}

We first showcase the accuracy, resource efficiency, and processing latency advantages of Slim-SC compared to the baseline methods. We then demonstrate the robustness of the approach with Random Pruning (RP) and Diversity-based Pruning (DP) and contrast our informed pruning strategy with naïve alternatives.

\subsection{Performance Comparison with Baselines}
\label{sec: effectiveness-slimsc-vs-baselines}
We summarize all the experimental results for Slim-SC and the baseline methods in Table~\ref{tab:accuracy-comparison}. All reported metrics are averaged over three independent runs to ensure stability.

\begin{table*}
\small
  \setlength{\tabcolsep}{3.9pt}
  \centering
  \begin{tabular}{l@{\hskip 1pt}|ccc|ccc|ccc|ccc}
    \hline
    \multicolumn{1}{c|}{\textbf{Methods}} & \multicolumn{12}{c}{\textbf{Datasets \& Metrics}} \\
    \hline
    & \textbf{D1} & \textbf{D2} & \textbf{D3} & \textbf{D1} & \textbf{D2} & \textbf{D3} & \textbf{D1} & \textbf{D2} & \textbf{D3} & \textbf{D1} & \textbf{D2} & \textbf{D3} \\
    & \multicolumn{3}{c|}{Accuracy (\%)} & \multicolumn{3}{c|}{Latency (s)} & \multicolumn{3}{c|}{\shortstack{Avg. Tokens \\ (thousands)}} & \multicolumn{3}{c}{\shortstack{Mean\\ KVC usage (\%)}} \\
    \hline
    \multicolumn{13}{c}{R1-Distill} \\
    \hline
    CoT & 
    \smallpm{58.8}[2.3] & \smallpm{67.8}[9.6] & \smallpm{89.8}[0.4] & 
    \smallpm{112}[3] & \smallpm{172}[18] & \smallpm{38}[2] &
    7 & 10 & 2 &
    2 & 2 & 1 \\\hline
    SC & 
    \smallpm{63.0}[0.6] & \textbf{\smallpm{82.2}[1.9]} & \smallpm{90.6}[0.4] &
    \smallpm{536}[16] & \smallpm{942}[29] & \smallpm{61}[3] &
    \smallpm{421}[5] & \smallpm{618}[4] & 18 &
    \smallpm{47}[2] & \smallpm{56}[2] & 4 \\
    ESC &
    \smallpm{62.0}[0.8] & \smallpm{81.1}[1.9] & \smallpm{89.5}[0.5] &
    \smallpm{876}[17] & \smallpm{1542}[100] & \textbf{\smallpm{51}[1]} &
    \textbf{\smallpm{262}[3]} & \textbf{\smallpm{392}[22]} & \textbf{5} &
    \textbf{10} & \textbf{13} & \textbf{1} \\
    CCoT-SC & \smallpm{60.4}[1.5] & \smallpm{80.0}[3.3] & \smallpm{89.9}[0.2] &
    \smallpm{505}[32] & \smallpm{797}[64] & \smallpm{58}[2] &
    \smallpm{402}[6] & \smallpm{572}[7] & 17 &
    \smallpm{46}[3] & \smallpm{44}[15] & 3 \\\hline
    \textbf{Slim-SC (DP)} & 
    \smallpm{62.5}[1.1] & \textbf{\smallpm{82.2}[1.9]}  & \textbf{\smallpm{90.8}[0.2]} &
    \textbf{\smallpm{296}[84]} & \textbf{\smallpm{536}[167]} & \smallpm{54}[2] &
    \smallpm{278}[67] & \smallpm{418}[107] & \smallpm{16}[1] &
    \smallpm{35}[9] & \smallpm{46}[12] & 3\\
    \textbf{Slim-SC (RP)} &
    \textbf{\smallpm{63.5}[1.8]} & \textbf{\smallpm{82.2}[1.9]} & 90.2 &
    \smallpm{381}[126] & \smallpm{664}[139] & \smallpm{56}[1] &
    \smallpm{328}[77] & \smallpm{486}[96] & 17 &
    43 & \smallpm{49}[8] & 3 \\\hline
    
    \multicolumn{13}{c}{QwQ-32B} \\
    \hline
    CoT & \smallpm{64.4}[3.7] & \smallpm{76.7}[3.3] & \smallpm{90.6}[0.4] &
    \smallpm{157}[12] & \smallpm{300}[41] & \smallpm{59}[1] &
    8 & \smallpm{13}[1] & 3 & 
    2 & 3 & 1 \\\hline
    SC & \textbf{\smallpm{66.8}[1.3]} & 83.3 & \smallpm{91.5}[0.5] &
    \smallpm{312}[28] & \smallpm{479}[75] & \smallpm{92}[1] &
    \smallpm{119}[1] & 105 & \smallpm{24}[1] &
    16 & \smallpm{15}[1] & 4\\
    ESC & \smallpm{66.0}[1.1] & \smallpm{80.0} & \textbf{\smallpm{91.6}[0.2]} &
    \smallpm{315}[20] & \smallpm{574}[17] & \smallpm{91}[4] &
    \textbf{\smallpm{26}[2]} & \textbf{\smallpm{47}[2]} & \textbf{8} &
    \textbf{3} & \textbf{5} & \textbf{1}\\
    CCoT-SC &
    \smallpm{66.7}[1.8] & \textbf{\smallpm{84.4}[1.9]} & \smallpm{91.5}[0.2] &
    \smallpm{286}[27] & \smallpm{435}[65] & \smallpm{89}[5] &
    111 & \smallpm{99}[2] & 22 &
    15 & 15 & 3 \\\hline
    \textbf{Slim-SC (DP)} & \smallpm{64.6}[1.5] & \textbf{\smallpm{84.4}[1.9]} & \smallpm{91.5}[0.2] &
    \textbf{\smallpm{250}[9]} & \smallpm{442}[19] & \textbf{84} &
    \smallpm{95}[6] & \smallpm{104}[2] & \smallpm{20}[1] &
    \smallpm{12}[1] & 15 & 3\\
    \textbf{Slim-SC (RP)} & \smallpm{65.8}[1.3] & \smallpm{82.2}[1.9]  & \textbf{\smallpm{91.6}[0.6]} &
    \smallpm{276}[15] & \textbf{\smallpm{431}[17]} & \textbf{\smallpm{84}[2]} &
    \smallpm{109}[11] & \smallpm{97}[6] &\smallpm{20}[2] &
    \smallpm{15}[1] & \smallpm{14}[1] & 3\\
    \hline
  \end{tabular}

  \caption{
    Accuracy, mean processing latency per question in seconds, mean number of completion tokens per question, and mean Key-Value Cache usage per question for baselines and Slim-SC on three datasets. \\
    \textbf{Dataset legend}: GPQA Diamond as D1, AIME-2024 as D2, AQuA-RAT as D3. \\
    \textbf{Result highlights:} Slim-SC achieves the same or higher accuracy than all the baselines, except vs. SC with QwQ-32B for GPQA. Slim-SC also delivers much lower latency for GPQA and AIME. Slim-SC uses fewer tokens and memory compared to all methods, except ESC.
  }
  \label{tab:accuracy-comparison}
\end{table*}

\noindent\textbf{Slim-SC maintains or improves accuracy over baselines.}
Across all models and datasets, Slim-SC consistently achieves performance comparable to or better than standard SC. With R1-Distill, Slim-SC~(RP) improves accuracy on \texttt{GPQA Diamond} (+0.5 pp over SC) and matches SC on \texttt{AIME-2024} (82.2\%), while outperforming other efficiency-focused baselines like ESC and CCoT-SC. A similar trend holds for QwQ-32B, where Slim-SC variants are highly competitive with SC and notably outperform ESC on complex reasoning tasks (D1 and D2). We attribute Slim-SC’s strong performance to its progressive pruning strategy, which, as shown in Appendix~\ref{appendix:candidate_pool_quality}, enriches the candidate pool with a higher proportion of correct chains by selectively removing redundant, incorrect ones.

\noindent\textbf{Slim-SC features a superior latency profile}. 
As shown in Table~\ref{tab:accuracy-comparison}, with R1-Distill, Slim-SC~(DP) lowers the latency of SC by 45\% on \texttt{GPQA}, 43\% on \texttt{AIME}, and 11\% on \texttt{AQuA}. With QwQ-32B, Slim-SC~(DP) trims the latency on \texttt{GPQA} by 20\%, on \texttt{AIME} by 8\%, and on \texttt{AQuA} by 9\%. Slim-SC~(RP) achieves comparable gains. 
Slim-SC provides a clear advantage in inference speed while retaining or improving accuracy.

In contrast, other efficiency-focused baselines struggle with this trade-off. CCoT-SC provides minimal speed-up, while ESC's sequential design creates a severe latency bottleneck. On R1-Distill, ESC is \emph{over 2.9$\times$ slower} than Slim-SC (DP) on \texttt{GPQA}. This high latency translates to poor user-facing response times and prolonged GPU occupancy. 
Furthermore, ESC’s overly aggressive early stopping can truncate exploring diverse reasoning pathways prematurely, forming consensus toward incorrect answers, as evidenced by its lower accuracy on R1-Distill D3 (89.5\%) compared to the single-chain CoT baseline (89.8\%).

\noindent\textbf{Slim-SC effectively reduces token and memory-time cost of SC}. 
By terminating redundant chains early, Slim-SC reduces the total number of generated tokens and mean KV Cache compared to SC.  For example, on the complex \texttt{AIME} task with R1-Distill, Slim-SC~(DP) slashes token usage by 32\% compared to SC (418k vs. 618k) and reduces the mean KV Cache by 10pp compared to SC (46\% vs. 56\%). A similar trend of substantial token and mean KV Cache reduction is observed across all datasets and models. In many instances Slim-SC is more efficient than CCoT-SC in these two metrics.
While Table~\ref{tab:accuracy-comparison} shows that ESC consistently generates the fewest tokens and has the lowest Mean KV Cache, this raw efficiency comes at a severe price. As analyzed in detail in Appendix~\ref{sec:kvcache_efficiency_analysis}, Slim-SC's architecture results in a much lower overall GPU-time cost. By completing tasks significantly faster, it frees up valuable hardware resources far more quickly than ESC's slow, sequential process. ESC's token frugality is a direct consequence of its aggressive early stopping, a mechanism that not only incurs a debilitating latency penalty but also introduces the risk of performance degradation, as previously discussed. Therefore, for practical deployment in high-throughput or time-billed environments, Slim-SC's approach offers a more cost-effective and reliable use of computational resources.

In summary, Slim-SC often matches or surpasses SC in accuracy and outperforms efficiency-focused baselines like ESC and CCoT-SC in most settings. While ESC offers an alternative for extremely token-constrained scenarios, Slim-SC provides a more balanced and practical solution. It is the preferred method when low latency and high accuracy are critical.

\subsection{Robustness of Slim-SC}
We test how Slim-SC is robust to the choice of pruning rule, pruning delay, and similarity thresholds.
The threshold values of Slim-SC for Table~\ref{tab:accuracy-comparison} are presented in Table~\ref{tab:Slim-SC-DP} and Table~\ref{tab:Slim-SC-RP}.

\subsubsection{Similarity Pruning Algorithms}

\begin{figure}[hbt!]
  \includegraphics[width=\columnwidth]{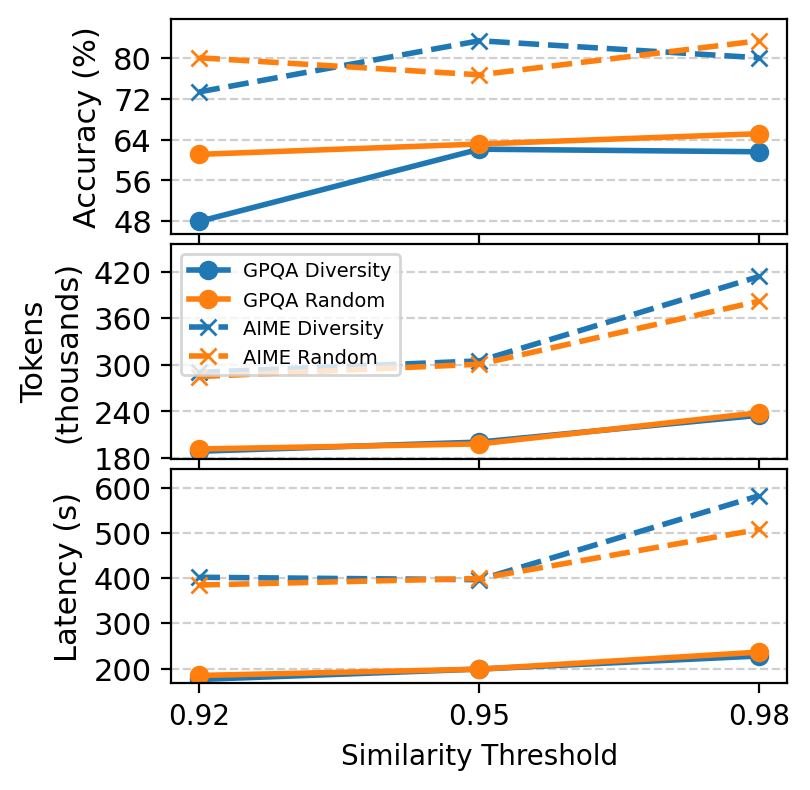}
  \caption{Accuracy, mean completion tokens per question, and processing latency of Slim-SC on R1-Distill when varying the similarity threshold for the diversity and random-based pruning methods in Slim-SC.}
  \label{fig:stats-threshold}
\end{figure}

\noindent\textbf{Random Selection is simple and effective while Internal Diversity delivers token efficiency}.
Our main experiments compare two selection heuristics for pruning a similar pair: Random Pruning (RP) and Diversity-based Pruning (DP). As shown in Table~\ref{tab:accuracy-comparison}, Slim-SC (RP) consistently delivers strong and stable accuracy, making it a simple, effective, and robust default strategy. Slim-SC (DP) offers an alternative trade-off; it often yields lower token usage by operating at a slightly lower similarity threshold (see Fig.~\ref{fig:stats-threshold}), which can be beneficial in resource-constrained settings, sometimes at the cost of a minor drop in accuracy.

\subsubsection{Ablation Study}

To validate that Slim-SC's performance stems from its informed, similarity-based strategy, we compare it against two na\"ive pruning baselines:
\begin{itemize}[leftmargin=*]
\item \textbf{Global Random Pruning}: At each analysis step, one active chain is randomly selected from the entire pool and pruned, without considering semantic content. This baseline tests whether deliberate pruning is necessary.  
\item \textbf{Least Similar Pruning}: At each step, the pair of thoughts from different chains with the lowest cosine similarity is identified. One of these two chains is then randomly pruned, actively removing diversity. This baseline tests whether pruning similar chains is indeed beneficial. 
\end{itemize}
As shown in Table~\ref{tab:pruning-ablation}, both strategies lead to a significant degradation in accuracy compared to Slim-SC. Least Similar Pruning is particularly harmful, with its accuracy on \texttt{AIME-2024} dropping to 74.4\%-a nearly 8pp loss compared to Slim-SC. This confirms that reasoning diversity is crucial for SC and that actively destroying it is detrimental. Global Random Pruning, while less harmful, is still clearly suboptimal as its unguided nature risks removing valuable chains. This ablation provides strong evidence that Slim-SC's performance gains are directly attributable to its core principle: intelligently targeting redundancy while preserving diversity.

\begin{table}[!t]
  \setlength{\tabcolsep}{6.5pt}
  \centering
  \begin{tabular}{l@{\hskip 1pt}|ccc}
    \hline
    \multicolumn{1}{c|}{\textbf{Methods}} & \multicolumn{3}{c}{\textbf{Datasets}} \\
    \hline
    & \textbf{D1} & \textbf{D2} & \textbf{D3} \\
    & \multicolumn{3}{c}{Accuracy (\%)} \\
    \hline
    CoT & \smallpm{64.4}[3.7] & \smallpm{76.7}[3.3] & \smallpm{90.6}[0.4] 
    \\\hline
    SC & \textbf{\smallpm{66.8}[1.3]} & 83.3 & \smallpm{91.5}[0.5] \\\hline
    GlobalRP & 
    \smallpm{64.1}[1.3] & \smallpm{77.8}[1.9] & \smallpm{91.2}[0.6] \\
    LeastSimP &
    \smallpm{62.6}[3.0] & \smallpm{74.4}[8.3] & \smallpm{91.1}[0.2] \\\hline
    \textbf{Slim-SC (DP)} & \smallpm{64.6}[1.5] & \textbf{\smallpm{84.4}[1.9]} & \smallpm{91.5}[0.2] \\
    \textbf{Slim-SC (RP)} & \smallpm{65.8}[1.3] & \smallpm{82.2}[1.9]  & \textbf{\smallpm{91.6}[0.6]} \\
    \hline
  \end{tabular}
  \caption{
    Compared to Slim-SC, accuracy drastically degrades when na\"ive pruning strategies like Global Random Pruning (GlobalRP) and Least Similar Pruning (LeastSimP) are used, when running the three datasets on QwQ-32B. \\
    \textbf{Dataset legend}: GPQA Diamond as D1, AIME-2024 as D2, AQuA-RAT as D3.
  }
  \label{tab:pruning-ablation}
\end{table}

\subsection{Hyperparameter Sensitivity Analysis}
\subsubsection{The Similarity Threshold }

\begin{figure}[hbt!]
  \includegraphics[width=\columnwidth]{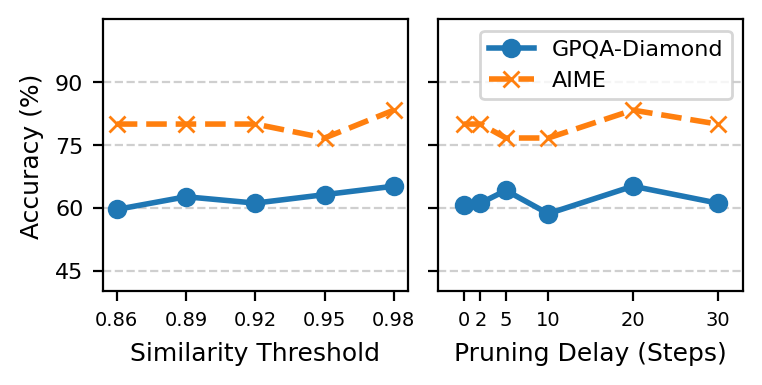}
  \caption{
  Accuracy remains stable for wide ranges of similarity thresholds and pruning delays when tested with R1-Distill. We choose the threshold of 0.98 and delay of 20 steps as more optimal.
  }
  \label{fig:acc-thresh-delay}
\end{figure}

\noindent\textbf{Performance is stable and robust at high similarity thresholds.}

As shown in Fig.~\ref{fig:acc-thresh-delay} (left), accuracy remains high and stable for $\tau_{sim} \geq 0.95$, typically peaking at $\tau_{sim}=0.98$. This empirical finding is not arbitrary; it is directly explained by the underlying similarity distributions of correct and incorrect chains, as detailed in Section \ref{sec:opp-for-improvement} and visualized in the appendix (\Cref{fig:pairwise-similarities}).

An analysis of these distribution plots reveals a crucial pattern. While incorrect chains are consistently more self-similar than correct ones across all thresholds, this disparity is often most pronounced around the $\tau_{sim}=0.98$ mark. For instance, on \texttt{AIME-2024} with R1-Distill (Fig. \ref{fig:motivation-insights}b), the proportion of incorrect-incorrect pairs among all similar pairs is maximized near this threshold, while the proportion of correct-correct pairs is minimized. This indicates that $\tau_{sim}=0.98$ is often the \emph{sweet spot} where our pruning strategy is maximally effective at its intended goal: disproportionately removing redundant incorrect chains while preserving the diversity among correct ones. This cleansing of the candidate pool provides a strong mechanistic justification for the peak in final accuracy observed at this threshold.

This insight leads to our practical recommendation: a high, fixed threshold like \emph{0.98 is not only a safe and robust choice but is also grounded in the observed clustering behavior of reasoning chains}. This allows practitioners to achieve optimal or near-optimal performance without needing to perform per-dataset tuning for this hyperparameter.

\subsubsection{Pruning Delay}

\noindent\textbf{Accuracy is stable across delay steps, peaking at a moderate delay of 20 steps.}
With each step defined as an interval of 3s, Fig. \ref{fig:acc-thresh-delay} (right) demonstrates the benefit of a pruning delay. While accuracy is relatively stable across various delay durations, it consistently peaks when the delay is set to 20 analysis steps for both \texttt{GPQA-Diamond} (63.5\%) and \texttt{AIME-2024} (82.2\%). This indicates that allowing chains a moderate period of 20 steps to develop their initial reasoning and diverge from one another is optimal before the similarity checks and pruning commence. This delay effectively preserves potentially correct and unique reasoning paths that might otherwise be pruned too early.

\section{Related Work}

\subsection{Test-Time Scaling (TTS)}
TTS refers to methods that enhance reasoning performance by investing additional computation at inference time, without modifying the model's parameters \cite{test-time-scaling-survey}. Existing TTS approaches can be categorized into sequential and parallel scaling. Sequential scaling extends the reasoning process by generating a series of intermediate steps before providing the final answer, as represented by Chain-of-Thought (CoT) prompting \cite{CoT}. The intermediate outputs can be further refined using methods like Self-Refine \cite{self-refine} or s1 \cite{s1}. However, the single-chain problem-solving trajectory can be compromised by error propagation, i.e., mistakes made in early steps may mislead subsequent reasoning.

Parallel scaling, by contrast, explores diverse reasoning paths simultaneously and selects the most plausible final answer based on certain criteria. Typical approaches include Best-of-N (BoN) sampling \cite{BoN} and SC decoding \cite{SC-CoT}. Both methods generate solutions in parallel, but differ in selection strategy: BoN uses a pretrained reward model to score the solutions based on its likelihood of yielding the correct answer \cite{reward-model}, whereas SC relies on majority voting among the predicted answers. Recent research explores latent reasoning with SC~\cite{xu2025softcottesttimescalingsoft}, further amplifying the benefit of both explicit and implicit reasoning~\cite{soft-cot} at test time. 
In summary, parallel scaling increases the chances of finding a correct solution and circumvents the potential error aggregation in sequential scaling. While it often enhances complex reasoning tasks, the effectiveness of TTS often requires non-trivial computational cost.

Besides TTS, system support that explores elastic inference scaling that right-sizes GPU cluster resources can improve resource efficiency while maintaining the accuracy \cite{fu2024serverlessllmlowlatencyserverlessinference, zhong2024distservedisaggregatingprefilldecoding, zhang2025blitzscalefastlivelarge}.

\subsection{Pruning Methods for Parallel Scaling}
Parallel scaling typically requires generating $N$ full reasoning paths,  each leading to a final answer. When $N$ is large enough, the performance gain can match that of post-training approaches \cite{fast-BoN}. However, generating all the full paths is computationally expensive, and many paths contribute little to the final decision. To address this, recent works focus on: 1) dynamically reducing $N$; and 2) halting low-quality paths early.

Instead of fixing the number of paths $N$ for all questions, methods such as Adaptive-Consistency (AC) \cite{ASC}, Early-Stopping Self-Consistency (ESC) \cite{ESC}, Difficulty-Adaptive Self-Consistency (DSC) \cite{DSC}, and Reasoning-Aware Self-Consistency (RASC) \cite{RASC} incrementally generate reasoning paths in small batches until the answers reach a consensus or a cap of $N$ is met. Specifically, ESC theoretically proves that this incremental sampling strategy maintains accuracy to a high degree. DSC measures question difficulty using an LLM to degenerate SC to single-chain CoT ($N=1$) for easy questions. RASC introduced a nuanced evaluation by incorporating both the answer consistency and reasoning quality via a weighted voting (similar to \citealp{confidence-SC}). Additionally, to accelerate finding the confident answer, Path-Consistency \cite{path-consistency-SC} reuses early reasoning steps from a confident path to guide the generation of subsequent paths. \emph{While effective, all these methods introduce latency due to sequential sampling, trading time for memory efficiency.}

Other approaches fix $N$ but halt the generation of unpromising paths early based on certain metrics. Fast BoN \cite{fast-BoN} uses the confidence scores from a reward model, whereas Self-Truncation BoN \cite{ST-BoN} introduces an internal consistency metric that computes the average squared distance between a path and all others. It keeps generating for \textit{a fixed time window $c$} and then prunes low-consistency paths. A similar strategy has also been recently applied in SC, i.e., Reasoning Pruning Perplexity Consistency (RPC) \cite{RPC}, which uses the model's own confidence to generate the correct answer, $p(y|x)$, after generating a reasoning step $x$. This paper contributes to this line by proposing a new efficient self-consistency method, Slim-SC, where we simply prune the chains based on the similarity of thoughts across the chains.

\section*{Conclusion}

This work introduces Slim-SC, a step-wise thought pruning method that improves the efficiency of SC without sacrificing accuracy. Motivated by the observation that correct and incorrect chains form distinct clusters, Slim-SC prunes highly similar chains to remove redundancy while preserving the diversity of reasoning paths.
Experiments show that Slim-SC substantially lowers 
latency and KVC usage by up to 45\% and 26\%, respectively, while
offering a more robust accuracy-latency trade-off than methods like ESC. 
Moreover, Slim-SC’s similarity-based pruning outperforms na\"ive pruning, confirming that preserving reasoning diversity is key to complex problem-solving. Finally, its simple Random Pruning (RP) variant makes Slim-SC a practical tool for deploying LLMs in latency-sensitive applications.

\section*{Limitations}

While Slim-SC demonstrates promise for improving SC-based efficiency, several limitations remain:  

\noindent\textbf{Metric Design:} Slim-SC uses similarity patterns to surface correct chains, but a more effective and lightweight metric is needed to fully bridge the gap between oracle and actual accuracy (Fig.~\ref{fig:oracle-actual-diff}).  

\noindent\textbf{Pruning vs. Reuse:} Our method prunes similar chains outright. Future work could merge partial reasoning segments or KV-cache states to improve both accuracy and computational reuse.  

\noindent\textbf{Scalability:} Evaluation should be extended to larger models and more diverse datasets to assess robustness and generalizability, as our study is currently limited to math reasoning with R1-Distill-14B and QwQ-32B.

\section*{Acknowledgments}

The authors thank the members of the HyScale lab at NTU Singapore for their constructive discussions and feedback on this work, and Vlad Pandelea for the guidance provided during the early stages of this work. This project is supported by the Ministry of Education, Singapore, under its Academic Research Fund Tier 1 (Project RS26/23). Xu Guo is supported by the Wallenberg-NTU Presidential Postdoctoral Fellowship and Colin Hong is supported by the A*STAR Graduate Scholarship.

\bibliography{custom}

\appendix

\begin{algorithm*}[h]
\caption{Slim-SC with Similarity Pruning}
\label{alg:slim_sc_pruning}
\begin{algorithmic}[1]
\State \textbf{Input:} Question $Q$, Initial number of chains $N$, Similarity threshold $\tau_{sim}$, Pruning strategy $\mathcal{P}_{strategy} \in \{\mathcal{P}_{rand}, \mathcal{P}_{div}\}$, Pruning delay parameters $D_{thoughts}, D_{steps}$.
\State \textbf{Output:} Final voted answer $A_{final}$.

\State Initialize $N$ chains $C = \{c_1, \dots, c_{N}\}$ generating thoughts for $Q$.
\State Initialize empty thought embedding index $F_{idx}$.
\State Initialize empty set of completed thoughts $T_{all} = \emptyset$.
\State $current\_step \gets 0$.

\While{any chain in $C$ is active AND $|C| > 1$}
    \State $current\_step \gets current\_step + 1$.
    \State Wait for new thought segments from active chains or analysis interval.
    \For{each active chain $c_i \in C$}
        \State Extract newly generated thought $th_{new}$ from $c_i$.
        \If{$th_{new}$ is not null}
            \State $e_{new} \gets \text{Embed}(th_{new})$.
            \State Add $e_{new}$ to $c_i$'s internal list of thought embeddings.
            \State $num\_thoughts_i \gets \text{count of thoughts in } c_i$.
            \If{$current\_step > D_{steps}$} \Comment{Pruning delay check}
                \State $(e_{neighbor}, c_{neighbor}) \gets \text{FindNearestNeighbor}(F_{idx}, e_{new}, \text{exclude\_chain}=c_i)$.
                \If{$c_{neighbor}$ is not null AND $\text{Similarity}(e_{new}, e_{neighbor}) > \tau_{sim}$}
                    \State $c_{prune} \gets \text{SelectChainToPrune}(c_i, c_{neighbor}, \mathcal{P}_{strategy})$ (Algorithm \ref{alg:select_prune_chain}).
                    \If{$|C| - |\{\text{chains marked for pruning}\}| > 1$}
                        \State Mark $c_{prune}$ as inactive.
                        \State Stop generation for $c_{prune}$.
                        \State Remove embeddings of $c_{prune}$ from $F_{idx}$.
                        \State \textbf{continue} to next chain or step.
                    \EndIf
                \EndIf
            \EndIf
            \State Add $(e_{new}, c_i, num\_thoughts_i)$ to $F_{idx}$.
        \EndIf
        \If{$c_i$ completes generation or reasoning phase ends for $c_i$}
            \State Mark $c_i$ so it's no longer checked for pruning but continues if not pruned.
        \EndIf
    \EndFor
    \If{all remaining active chains have completed reasoning}
        \State \textbf{break} \Comment{No more pruning possible}
    \EndIf
\EndWhile

\State Wait for all remaining active chains in $C$ to complete generation.
\State Collect final answers from non-pruned, completed chains.
\State $A_{final} \gets \text{MajorityVote}(\text{collected answers})$.
\State \Return $A_{final}$.
\end{algorithmic}
\end{algorithm*}

\begin{algorithm*}[t!]
\caption{SelectChainToPrune($c_a, c_b, \mathcal{P}_{strategy}$)}
\label{alg:select_prune_chain}
\begin{algorithmic}[1]
\State \textbf{Input:} Chains $c_a, c_b$; Pruning strategy $\mathcal{P}_{strategy}$.
\State \textbf{Output:} Chain selected for pruning.

\If{$\mathcal{P}_{strategy} = \mathcal{P}_{rand}$}
    \State \Return Randomly chosen chain from $\{c_a, c_b\}$.
\ElsIf{$\mathcal{P}_{strategy} = \mathcal{P}_{div}$}
    \State $S_{internal}(c_a) \gets \text{CalculateInternalDiversityScore}(c_a)$.
    \State $S_{internal}(c_b) \gets \text{CalculateInternalDiversityScore}(c_b)$.
    \If{$S_{internal}(c_a) > S_{internal}(c_b)$} \Comment{Higher score means less diverse}
        \State \Return $c_a$.
    \ElsIf{$S_{internal}(c_b) > S_{internal}(c_a)$}
        \State \Return $c_b$.
    \Else \Comment{Tie-breaking: internal diversity scores are equal.}
        \State $num\_thoughts_a \gets \text{count of thoughts in } c_a$.
        \State $num\_thoughts_b \gets \text{count of thoughts in } c_b$.
        \If{$num\_thoughts_a \le num\_thoughts_b$} \Comment{Heuristic: prune chain w/ fewer/eq thoughts.}
            \State \Return $c_a$.
        \Else
            \State \Return $c_b$.
        \EndIf
    \EndIf
\EndIf
\Procedure{CalculateInternalDiversityScore}{$c$}
    \State $E_c \gets \text{list of thought embeddings for chain } c$.
    \If{$|E_c| < 2$} \Return $0.0$. \EndIf
    \State $sum\_sim \gets 0$; $pair\_count \gets 0$.
    \For{$i \gets 0$ to $|E_c|-2$}
        \For{$j \gets i+1$ to $|E_c|-1$}
            \State $sum\_sim \gets sum\_sim + \text{Similarity}(E_c[i], E_c[j])$.
            \State $pair\_count \gets pair\_count + 1$.
        \EndFor
    \EndFor
    \State \Return $sum\_sim / pair\_count$ if $pair\_count > 0$ else $0.0$.
\EndProcedure
\end{algorithmic}
\end{algorithm*}

\section{Analysis of Self-Consistency}
\label{sec:analysis_of_sc}

\subsection{Divergence in the Mean Completion Tokens}\label{appendix:completion-tokens-cdf}

\begin{figure}[H]
\centering
\includegraphics[width=\linewidth]{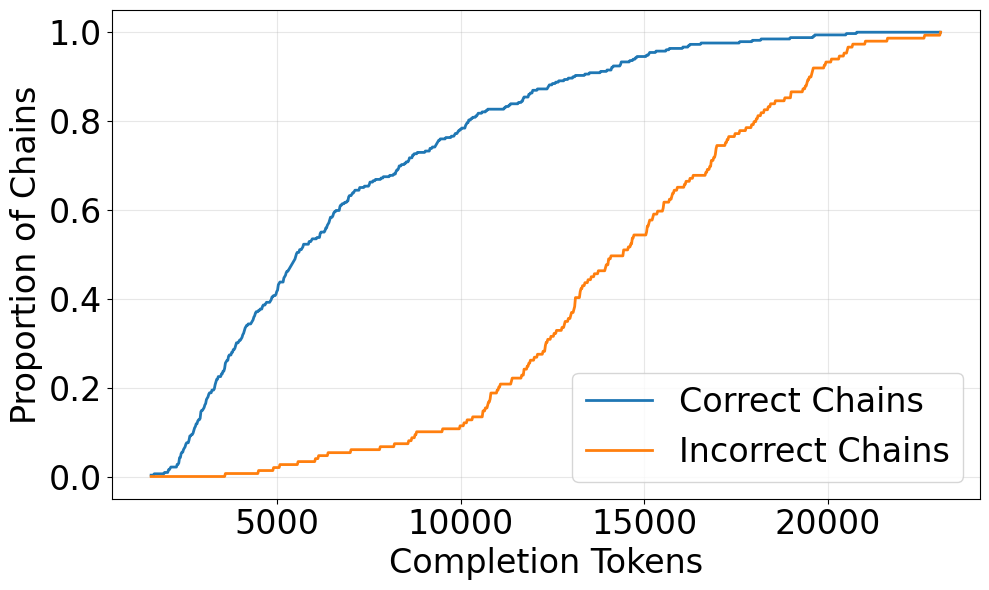}
\caption{CDF of chains by completion tokens using R1-Distill on AIME with N=16}
\label{fig:completion-tokens-cdf}
\end{figure}

\begin{figure*}[ht]
\begin{subfigure}[b]{0.32\textwidth}
  \includegraphics[width=\columnwidth]{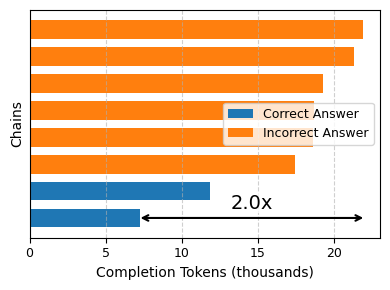}
  \caption{SC ($N = 8$) for R1-Distill on AIME-2024 Q27.
  }
\end{subfigure}
\hfill
\begin{subfigure}[b]{0.33\textwidth}
\includegraphics[width=\columnwidth]{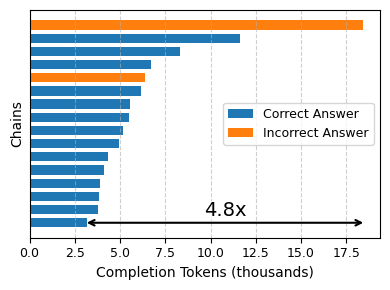}
\caption{SC ($N = 16$) for R1-Distill on AIME-2024 Q29.
}
\end{subfigure}
\hfill
\begin{subfigure}[b]{0.32\textwidth}
\includegraphics[width=\columnwidth]{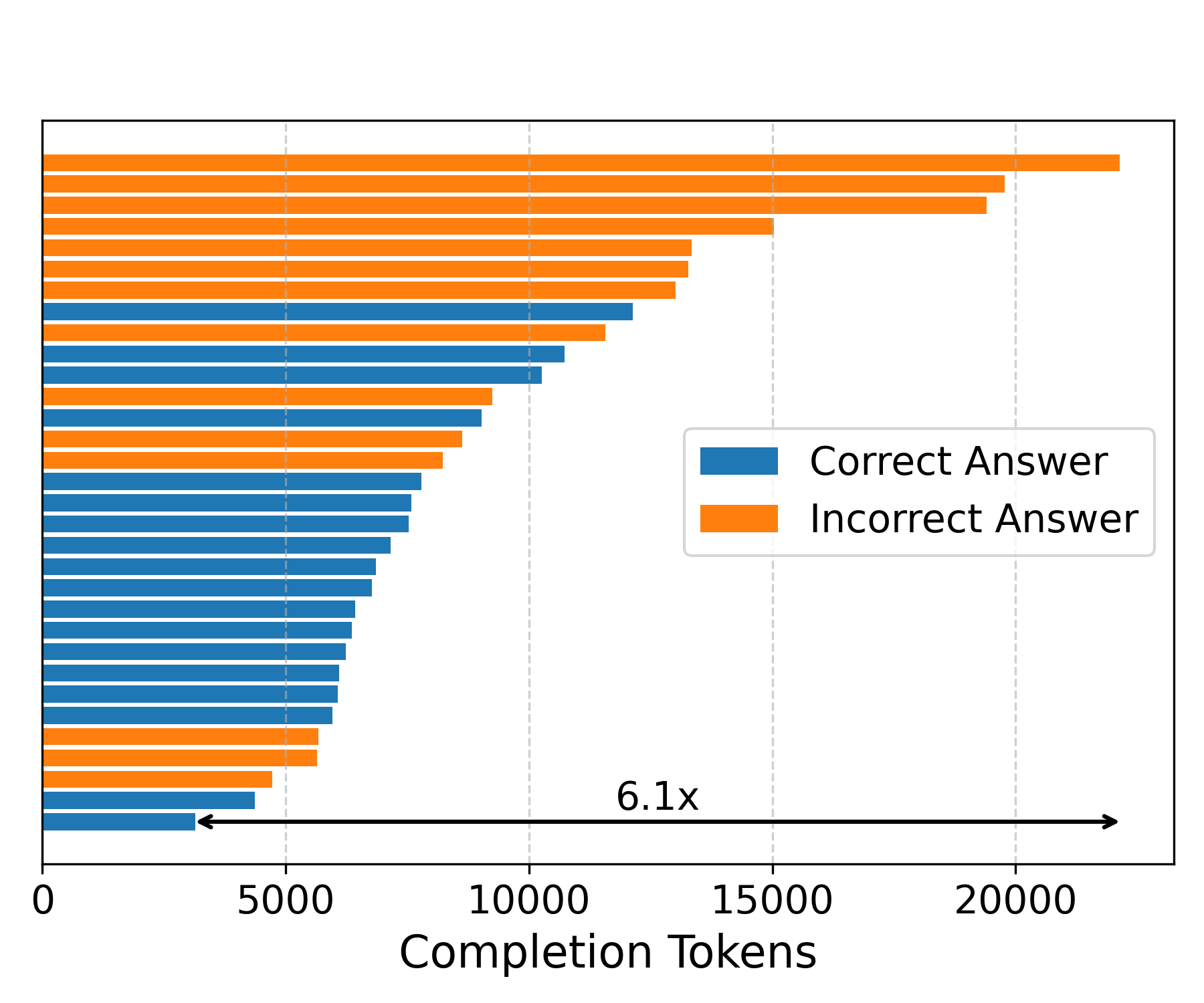}
\caption{SC ($N = 32$) for R1-Distill on AIME-2024 Q29.
}
\end{subfigure}
\caption{SC \emph{wait-for-all} effect at other $N$s.}
\label{fig:wait-for-all}
\end{figure*}

\begin{figure*}[h!]
    \centering
    \begin{subfigure}[t]{0.48\textwidth}
        \centering
        \includegraphics[width=\linewidth]{figures/stacked_bar_plot_granular_r1_gpqa.png}
        \caption{Pairwise similarity on \texttt{GPQA-Diamond} (R1-Distill).}
        \vspace{1em}
        
        \includegraphics[width=\linewidth]{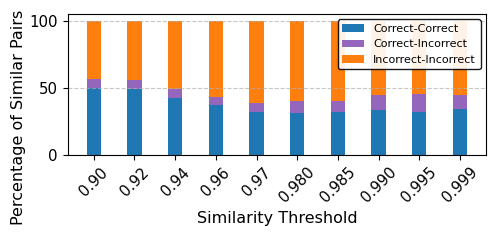}
        \caption{Pairwise similarity on \texttt{AQuA-RAT} (R1-Distill).}
    \end{subfigure}
    \hfill
    \begin{subfigure}[t]{0.48\textwidth}
        \centering
        \includegraphics[width=\linewidth]{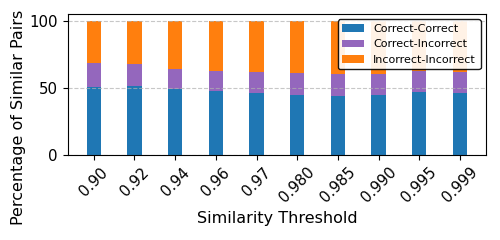}
        \caption{Pairwise similarity on \texttt{GPQA-Diamond} (QwQ).}
        \vspace{1em}
        
        \includegraphics[width=\linewidth]{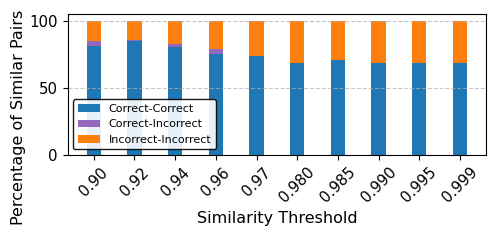}
        \caption{Pairwise similarity on \texttt{AIME-2024} (QwQ).}
        \vspace{1em}
        
        \includegraphics[width=\linewidth]{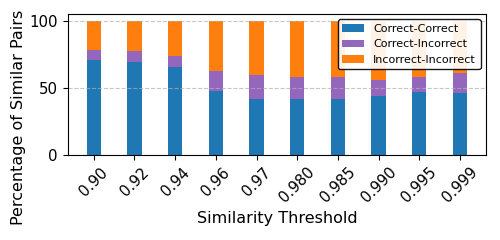}
        \caption{Pairwise similarity on \texttt{AQuA-RAT} (QwQ).}
    \end{subfigure}
    
    \caption{Pairwise similarities using R1-Distill (left) and QwQ (right) across various datasets.}
    \label{fig:pairwise-similarities}
\end{figure*}

Our analysis in Section \ref{problems-scaling-sc} highlights that standard SC often waits for long, incorrect chains. This observation is further substantiated by examining the distribution of completion tokens for correct versus incorrect chains. Fig. \ref{fig:completion-tokens-cdf} presents the Cumulative Distribution Function (CDF) of completion tokens for chains generated by R1-Distill on the \texttt{AIME} dataset (with $N=16$).

The CDF reveals a distinct divergence:
\begin{itemize}
    \item \textbf{Correct chains tend to be shorter.} A significant proportion of correct chains complete with fewer tokens. For instance, approximately 60\% of correct chains are completed within 7,500 tokens.
    \item \textbf{Incorrect chains are generally longer.} The CDF for incorrect chains rises much more slowly, indicating that a larger number of tokens is typically required before an incorrect chain completes. Only about 20\% of incorrect chains are completed within 10,000 tokens, whereas nearly 80\% of correct chains have finished by this point.
\end{itemize}
This divergence in completion token lengths, where incorrect chains are frequently longer, reinforces the inefficiency of the \emph{wait-for-all} SC strategy. It also supports the rationale behind pruning, as selectively removing chains (especially those that are both incorrect and lengthy, or highly similar to other incorrect, lengthy chains as suggested by Fig. \ref{fig:similar-thoughts-lead-to-same-answer}) could save substantial computational resources without necessarily compromising the chances of finding a correct answer among the shorter, more efficient correct chains. This observed difference in token usage profiles further motivates strategies like Slim-SC that aim to curtail the generation of overly long or redundant paths.

\subsection{\emph{Wait-for-all} effect in SC}
Fig. \ref{fig:wait-for-all} demonstrates that the \emph{wait-for-all} problem is not only experienced at $N=64$ as shown in Fig. \ref{fig:sc-problems}c, but experienced to an increasing degree as $N$ increases.
\label{appendix:wait-for-all}

\subsection{Categorizing similar chains}\label{appendix:prop_of_sim_chain}
To produce the results in Fig.~\ref{fig:motivation-insights}b, we begin with the textual outputs from our SC ($N=64$) experiment on R1-Distill using the GPQA Diamond dataset. We then apply a modified version of Algorithm~\ref{alg:slim_sc_pruning}. Specifically, we define a step as 100 tokens, based on our experimental estimate of the average number of tokens generated per step in the final setup. We set $D_{\text{steps}} = 20$ and proceed with the algorithm, except that instead of performing the pruning steps (lines 19–23), we record all chain pairs that were deemed similar and classify each pair based on whether both chains were correct, both incorrect, or one correct and one incorrect. We include similar charts for the rest of the models and datasets at their corresponding ideal $N$ (from Table~\ref{tab:optimal-n}) in \Cref{fig:pairwise-similarities}.

\subsection{Impact of Pruning on the Quality of the Candidate Pool}
\label{appendix:candidate_pool_quality}

Our core motivation, as discussed in Section~\ref{sec:opp-for-improvement}, is that incorrect reasoning chains tend to form denser semantic clusters than correct ones. While Fig.~\ref{fig:similar-thoughts-lead-to-same-answer} illustrates this by showing the composition of similar pairs, it does not directly demonstrate that our pruning method improves the quality of the final candidate set used for voting.

To provide this direct evidence, we analyzed the average proportion of correct chains remaining in the candidate pool after pruning, compared to the initial proportion in standard SC. The results are presented in Table~\ref{tab:correct_chain_prop_r1} and Table~\ref{tab:correct_chain_prop_qwq}.

The results in these tables confirm our hypothesis. Across both models, \textbf{Slim-SC (RP) increases the proportion of correct chains in 5 out of 6 experimental settings} compared to standard SC. This provides strong, direct evidence that our similarity-based pruning strategy is working as intended: it preferentially removes redundant incorrect chains more often than correct ones. This \emph{cleansing} of the candidate pool mechanistically explains why Slim-SC can maintain or even improve upon the final accuracy of SC, despite using fewer computational resources.

\begin{table}[h!]
\centering
\begin{tabular}{l|ccc}
\hline
\textbf{Method}       & \textbf{D1} & \textbf{D2} & \textbf{D3} \\ \hline
SC             & 0.578        & 0.717        & 0.879        \\
Slim-SC (DP)   & 0.585       & 0.716       & \textbf{0.905}        \\
Slim-SC (RP)   & \textbf{0.599} & \textbf{0.735} & 0.900        \\ \hline
\end{tabular}
\caption{R1-Distill-Qwen-14B: Average proportion of correct chains in the final candidate pool.}
\label{tab:correct_chain_prop_r1}
\end{table}

\begin{table}[h!]
\centering
\begin{tabular}{l|ccc}
\hline
\textbf{Method}       & \textbf{D1} & \textbf{D2} & \textbf{D3} \\ \hline
SC             & 0.637        & 0.775        & 0.916        \\
Slim-SC (DP)   & 0.626       & 0.767       & 0.915        \\
Slim-SC (RP)   & \textbf{0.657} & 0.773       & \textbf{0.917}        \\ \hline
\end{tabular}
\caption{QwQ-32B: Average proportion of correct chains in the final candidate pool.}
\label{tab:correct_chain_prop_qwq}
\end{table}

\subsection{Understanding KV Cache Efficiency: Peak vs. Duration}
\label{sec:kvcache_efficiency_analysis}

\begin{table}[h]
  \setlength{\tabcolsep}{14pt}
  \centering
  \begin{tabular}{l@{\hskip 1pt}|ccc}
    \hline
    \multicolumn{1}{c|}{\textbf{Methods}} & \multicolumn{3}{c}{\textbf{Datasets}} \\
    \hline
    & \textbf{D1} & \textbf{D2} & \textbf{D3} \\
    & \multicolumn{3}{c}{Mean Peak KVCache (\%)} \\
    \hline
    \multicolumn{4}{c}{R1-Distill} \\
    \hline
    CoT & 
    3 & 5 & 1
    \\\hline
    SC & 
    86 & \smallpm{94}[2] & \smallpm{8}[2] \\
    ESC &
    21 & \smallpm{28}[1] & 2 \\
    CCoT-SC & 84 & \smallpm{81}[15] & 6 \\\hline
    Slim-SC (DP) & 
    82 & \smallpm{92}[2] & 6 \\
    Slim-SC (RP) &
    85 & 93 & 6 \\\hline
    
    \multicolumn{4}{c}{QwQ-32B} \\
    \hline
    CoT & 
    3 & 5 & 1 \\\hline
    SC & 
    31 & \smallpm{30}[1] & 7\\
    ESC & 
    5 & 9 & 2\\
    CCoT-SC &
    29 & 28 & 6 \\\hline
    Slim-SC (DP) & 
    \smallpm{25}[2] & \smallpm{29}[1] & 6\\
    Slim-SC (RP) & 
    \smallpm{29}[2] & \smallpm{27}[2] & 6\\
    \hline
  \end{tabular}
  \caption{
    Mean Peak Key-Value Cache usage per question for baselines and Slim-SC on three datasets. \\
    \textbf{Dataset legend}: GPQA Diamond as D1, AIME 2024 as D2, AQuA-RAT as D3.
  }
  \label{tab:mean_peak_kvc}
\end{table}

While Table~\ref{tab:mean_peak_kvc} reports Mean Peak KV Cache usage, this single metric can be misleading if not considered in the context of processing time. A more holistic measure of resource cost is the \emph{GPU-time product}, which reflects the total duration a hardware accelerator is occupied. For time-billed cloud instances or high-throughput on-premise clusters, minimizing the total time a GPU is reserved for a single request is paramount for both cost-efficiency and serving capacity.

As shown in Table~\ref{tab:mean_peak_kvc}, ESC achieves the lowest Mean Peak KV Cache usage. However, its sequential nature leads to prolonged GPU occupancy, as it processes small batches over a much longer duration (see Latency in Table~\ref{tab:accuracy-comparison}). This results in a high overall GPU-time cost.

\begin{figure*}[ht]
    \centering
    \begin{subfigure}[t]{0.48\textwidth}
        \centering
        \includegraphics[width=\linewidth]{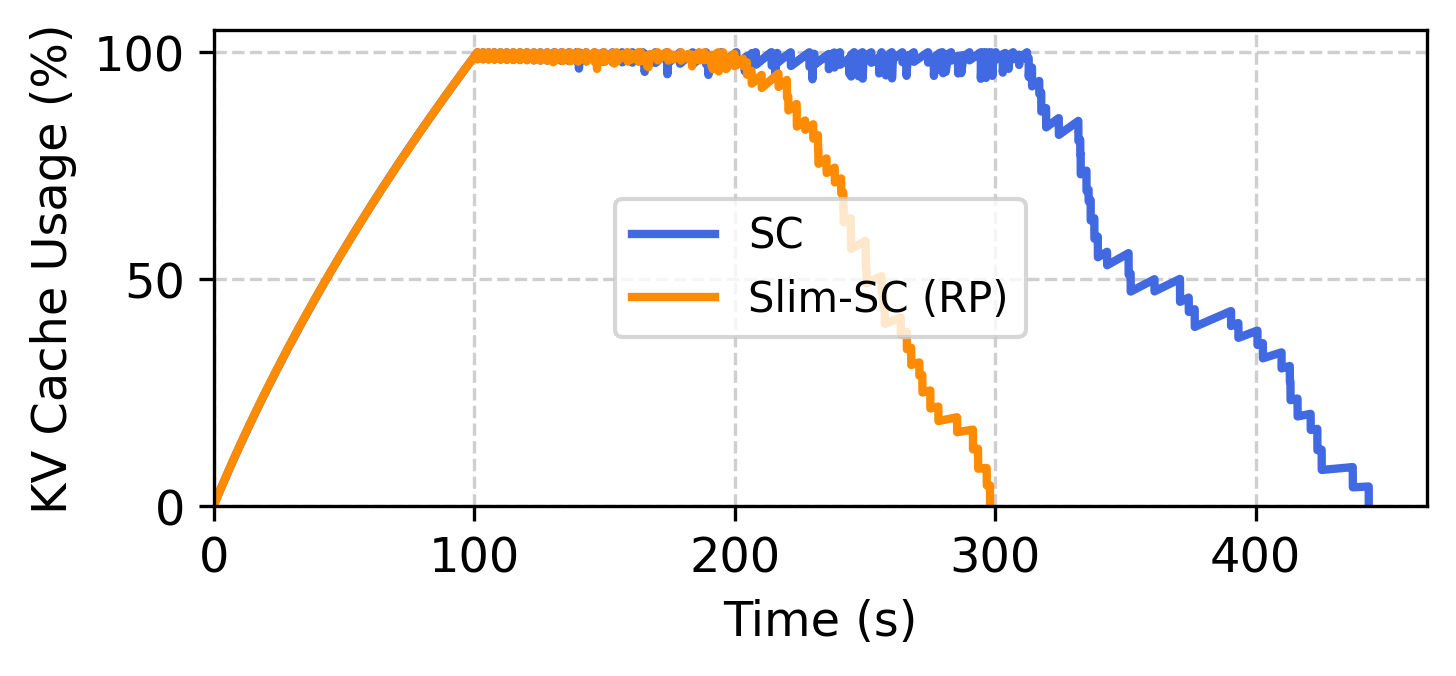}
        \caption{GPQA-Diamond Q2 (R1-Distill).}
        \vspace{1em}
        
        \includegraphics[width=\linewidth]{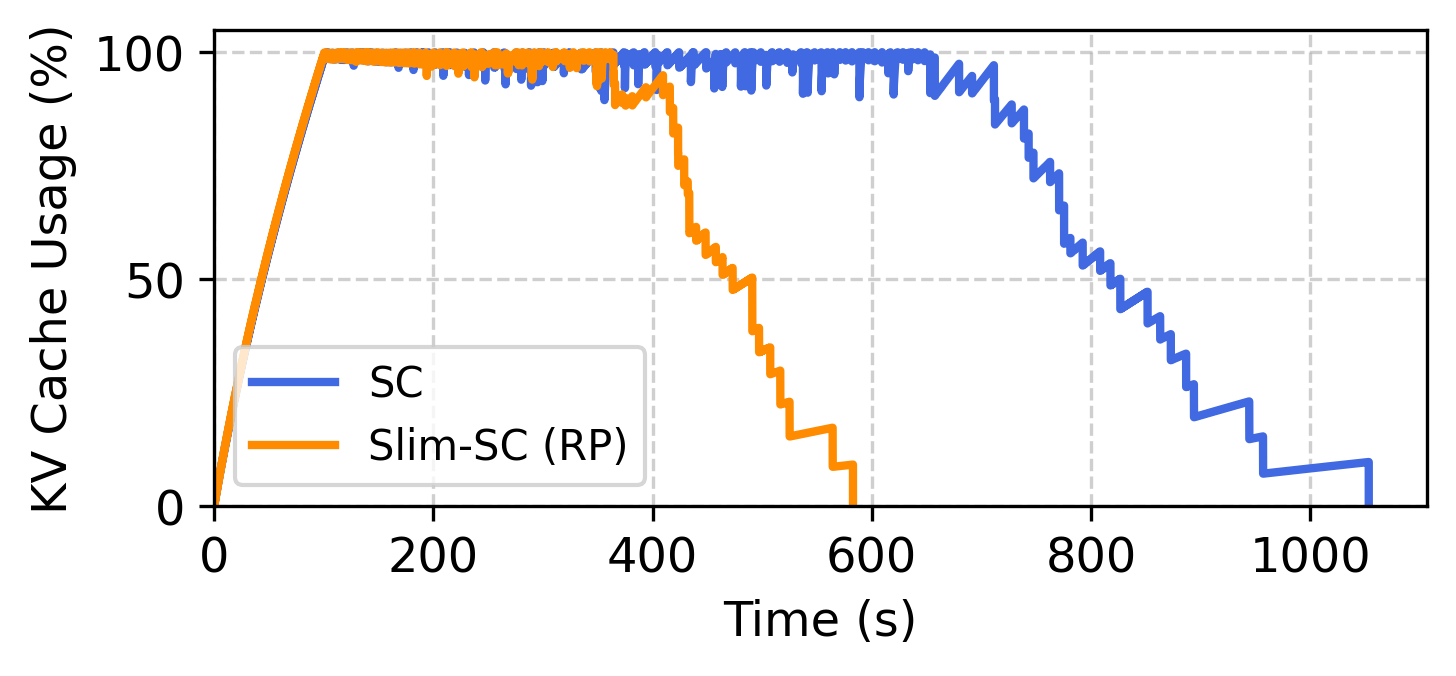}
        \caption{AIME-2024 Q22 (R1-Distill).}
        \vspace{1em}
        
        \includegraphics[width=\linewidth]{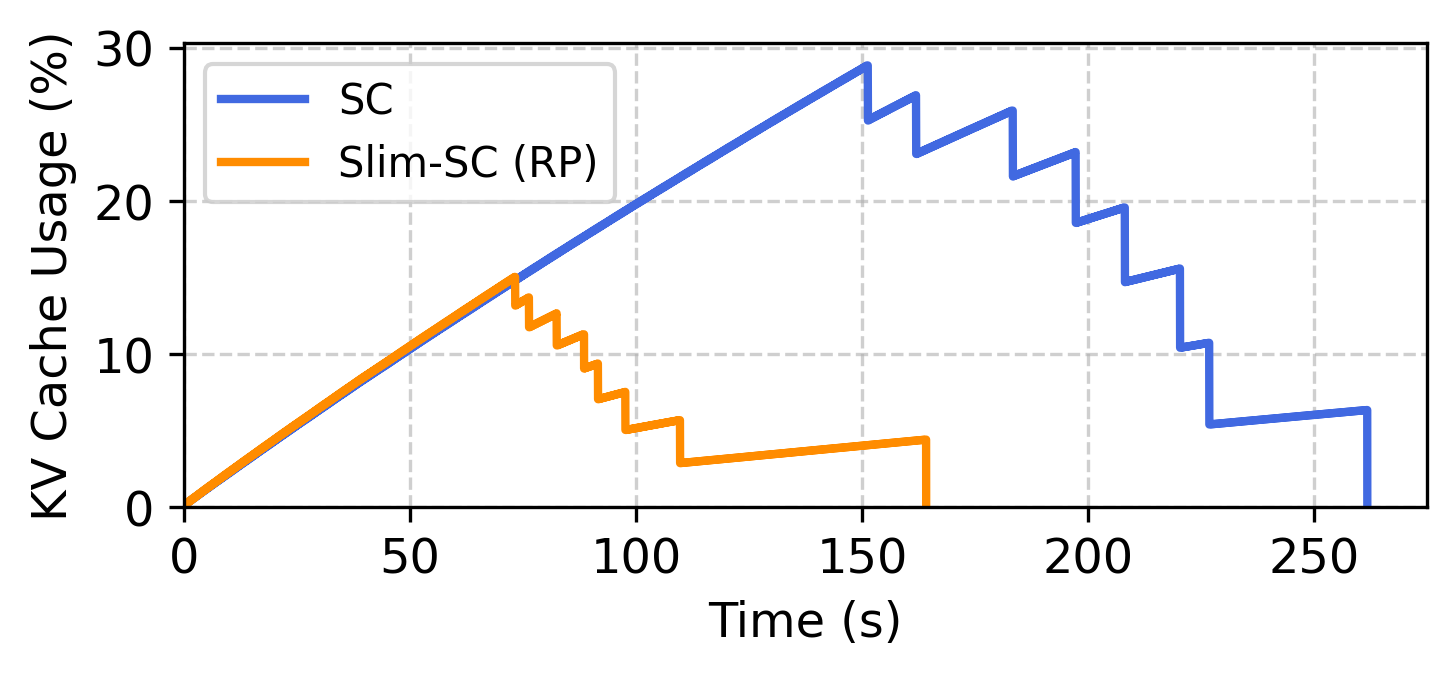}
        \caption{AQuA-RAT Q70 (R1-Distill).}
    \end{subfigure}
    \hfill
    \begin{subfigure}[t]{0.48\textwidth}
        \centering
        \includegraphics[width=\linewidth]{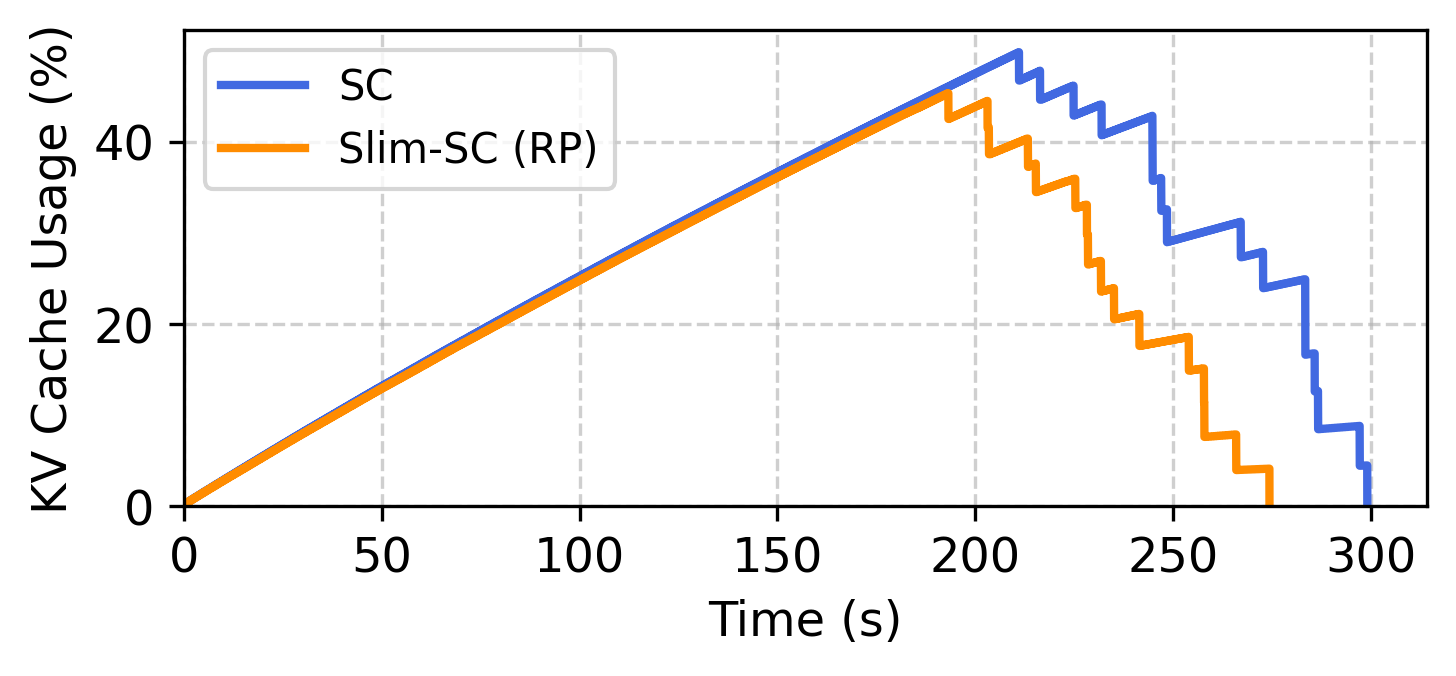}
        \caption{GPQA-Diamond Q70 (QwQ).}
        \vspace{1em}
        
        \includegraphics[width=\linewidth]{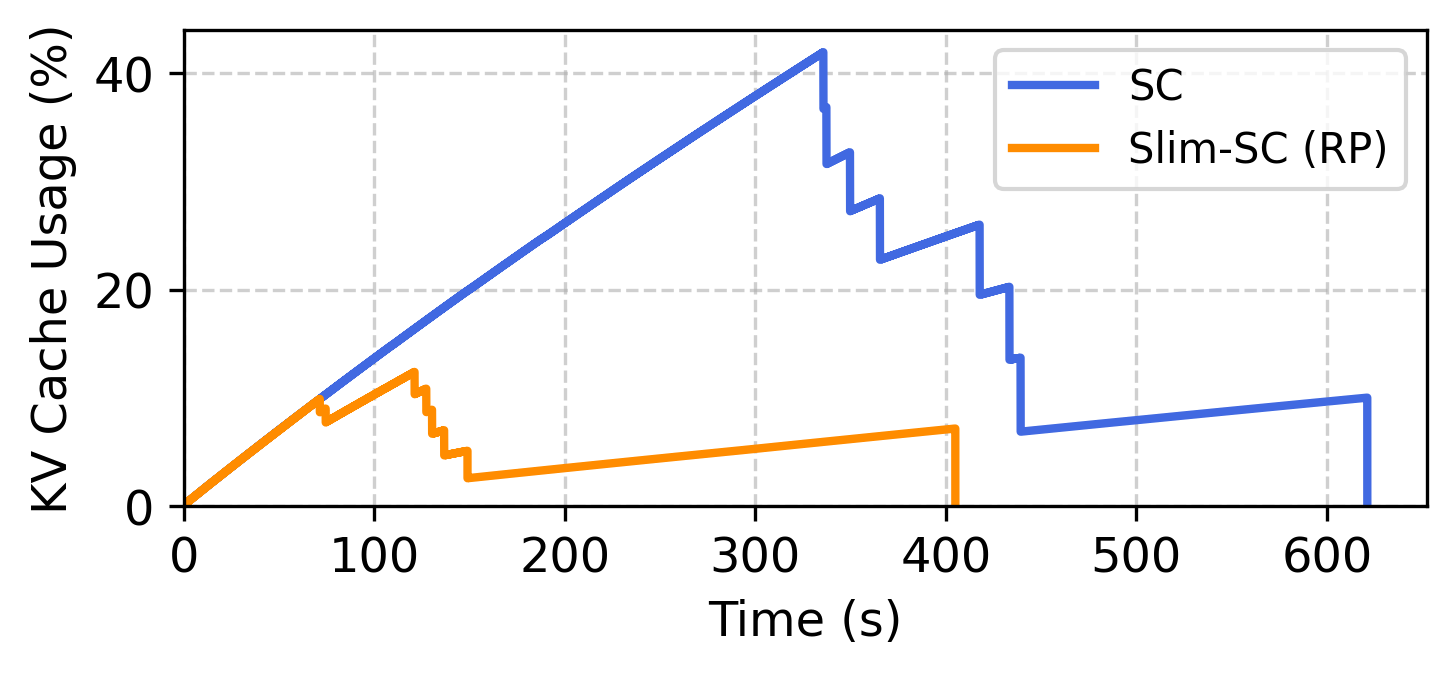}
        \caption{AIME-2024 Q8 (QwQ).}
        \vspace{1em}
        
        \includegraphics[width=\linewidth]{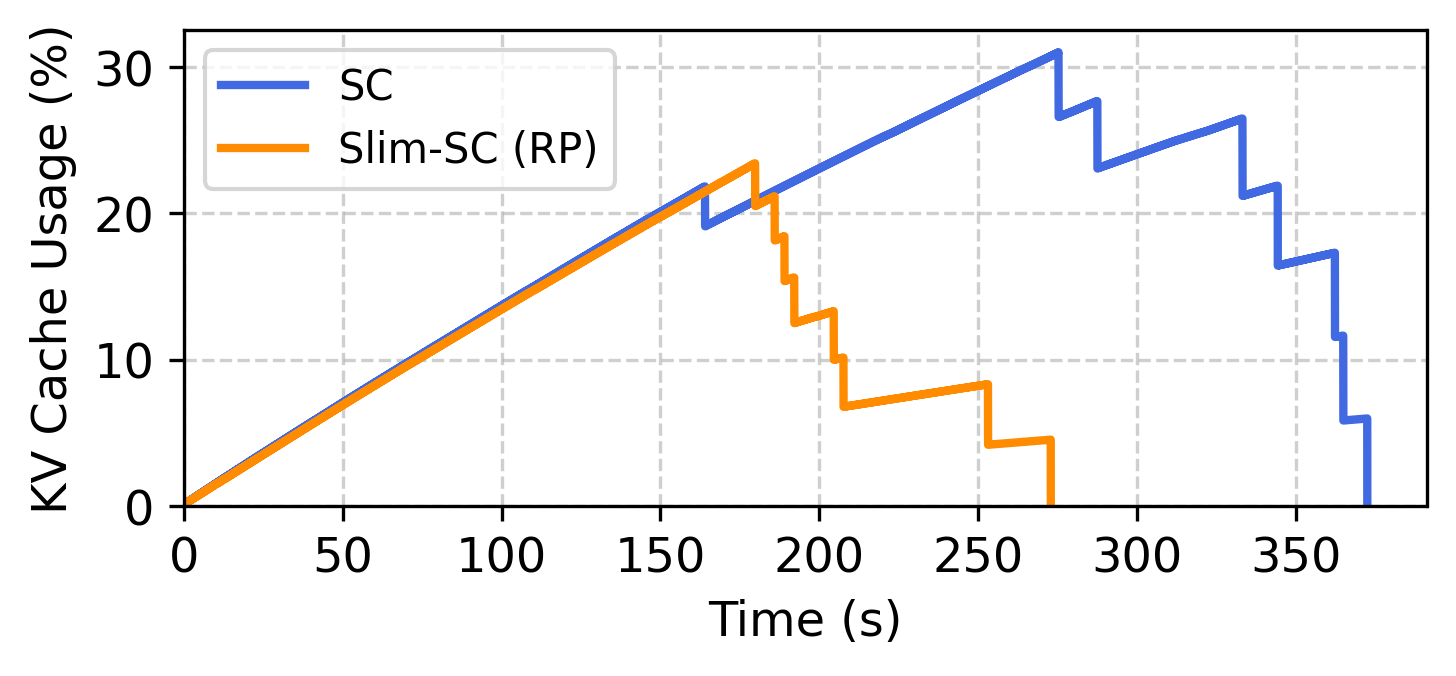}
        \caption{AQuA-RAT Q242 (QwQ).}
    \end{subfigure}
    
    \caption{Comparison of KVCache usage (\%) over time between SC and Slim-SC (RP) for R1-Distill (left) and QwQ (right).}
    \label{fig:kvcache-time-combined}
\end{figure*}

In contrast, Fig.~\ref{fig:kvcache-time-combined} illustrates the superior resource-time profile of Slim-SC.
\begin{itemize}
    \item \textbf{Initial Peak:} Slim-SC launches all chains in parallel, causing its KV Cache usage to momentarily peak at a level similar to standard SC. This initial ramp-up explains the high \emph{Mean Peak} values in the table.
    \item \textbf{Rapid Decline:} Critically, once the pruning mechanism activates, the KV cache usage for Slim-SC (orange line) rapidly plummets. The resources are freed up much faster as redundant chains are terminated.
    \item \textbf{Sustained Cost of Baselines:} Standard SC (blue line) exhibits sustained high KV Cache usage for a much longer period until chains naturally complete. ESC, while not plotted, would show a sawtooth pattern of smaller peaks over an even longer timeframe, occupying the GPU for the entire extended latency period.
\end{itemize}

Slim-SC's strategy of utilizing higher memory for a short burst is significantly more efficient in terms of total GPU-time cost than ESC's strategy of using low memory over a prolonged period. By finishing tasks faster, Slim-SC frees up valuable GPU resources for subsequent requests, making it a more practical and cost-effective solution for real-world, latency-sensitive applications.

\section{Experimental Details}
\label{sec:experimental_details}

\subsection{ESC Reproduction Details.}\label{appendix:ESC-reproduction}
Following the original work's setup for $N_{max}=40$ (where they used $W=5$, i.e., 8 windows), we aim for a similar number of maximum windows. Thus, we set $W = \max(2, \lceil N / 8 \rceil)$. The $\max(2, \dots)$ ensures that $W \ge 2$ to avoid degenerating to simple CoT when $N$ is small. For instance, if $N=64$, $W=8$. If $N=8$, $W=2$. The optimal setting searched for ESC is presented in Table \ref{tab:ESC-param}.

\begin{table}[h]
    \centering
    \begin{tabular}{c|c|c|c}
      \hline
      \textbf{Model} & \textbf{GPQA D.} & \textbf{AIME'24} & \textbf{AQuA-R.} \\
    \hline
     R1-Distill & 8 & 8 & 2 \\\hline
     QwQ-32B & 2 & 2 & 2\\\hline
    \end{tabular}
    \caption{The window size parameter searched for ESC.
    }
    \label{tab:ESC-param}
\end{table}

\subsection{Prompt Templates}

To ensure standardization, the following are the templates for the prompts we used for the \texttt{GPQA-Diamond}, \texttt{AIME-2024} and  \texttt{AQuA-RAT} benchmarks.

\begin{tcolorbox}[title=CoT/SC (AIME-2024), colback=gray!5!white, colframe=black!75!black]
\textbf{Prompt:} \\
Answer the following math problem.\\The last line of your response should be your integer answer within \\boxed\{\{\}\}.\\\{question\_text\}\\Put your final answer within \\boxed\{\{\}\}\\Think step by step before answering.
\vspace{0.5em}
\end{tcolorbox}

\begin{tcolorbox}[title=CCoT (AIME-2024), colback=gray!5!white, colframe=black!75!black]
\textbf{Prompt:} \\
Answer the following math problem.\\The last line of your response should be your integer answer within \\boxed\{\{\}\}.\\\{question\_text\}\\Put your final answer within \\boxed\{\{\}\}\\Think step by step before answering. Be concise.
\vspace{0.5em}
\end{tcolorbox}

\begin{tcolorbox}[title=CoT/SC (AQuA-RAT), colback=gray!5!white, colframe=black!75!black]
\textbf{Prompt:} \\
Answer the following multiple-choice question.
Think step-by-step to reach the solution.
Conclude your response with a single line containing the answer letter formatted exactly as 'Answer: \$LETTER'.

Question: \{question\_text\}

Options:
\{options\}
\vspace{0.5em}
\end{tcolorbox}

\begin{tcolorbox}[title=CCoT (AQuA-RAT), colback=gray!5!white, colframe=black!75!black]
\textbf{Prompt:} \\
Answer the following multiple-choice question.
Think step-by-step to reach the solution. Be concise.
Conclude your response with a single line containing the answer letter formatted exactly as 'Answer: \$LETTER'.

Question: \{question\_text\}

Options:
\{options\}
\vspace{0.5em}
\end{tcolorbox}

\begin{tcolorbox}[title=CoT/SC (GPQA-Diamond), colback=gray!5!white, colframe=black!75!black]
\textbf{Prompt:} \\
Answer the following multiple-choice science question.
Think step-by-step to reach the solution.
Conclude your response with a single line containing the answer letter formatted exactly as 'Answer: \$LETTER'.

Question: \{question\_text\}

Options: \{options\}
\vspace{0.5em}
\end{tcolorbox}

\begin{tcolorbox}[title=CCoT (GPQA-Diamond), colback=gray!5!white, colframe=black!75!black]
\textbf{Prompt:} \\
Answer the following multiple-choice science question.
Think step-by-step to reach the solution. Be concise.
Conclude your response with a single line containing the answer letter formatted exactly as 'Answer: \$LETTER'.

Question: \{question\_text\}

Options: \{options\}
\vspace{0.5em}
\end{tcolorbox}

\subsection{Optimal $N$ for standard SC}

\begin{table}[H]
    \centering
    \begin{tabular}{c|c|c|c}
      \hline
      \textbf{Model} & \textbf{GPQA D.} & \textbf{AIME'24} & \textbf{AQuA-R.} \\
    \hline
     R1-Distill & 64 & 64 & 8 \\\hline
     QwQ-32B & 16 & 8 & 8 \\\hline
    \end{tabular}
    \caption{The optimal N searched for the baseline Self-Consistency on each dataset and model.
    }
    \label{tab:optimal-n}
\end{table}

\subsection{Optimal thresholds for Slim-SC}

\begin{table}[H]
    \centering
    \begin{tabular}{c|c|c|c}
      \hline
      \textbf{Model} & \textbf{GPQA D.} & \textbf{AIME'24} & \textbf{AQuA-R.} \\
    \hline
     R1-Distill & 0.95 & 0.95 & 0.95 \\\hline
     QwQ-32B & 0.95 & 0.98 & 0.98 \\\hline
    \end{tabular}
    \caption{The similarity threshold empirically chosen for Slim-SC (DP).
    }
    \label{tab:Slim-SC-DP}
\end{table}

\begin{table}[H]
    \centering
    \begin{tabular}{c|c|c|c}
      \hline
      \textbf{Model} & \textbf{GPQA D.} & \textbf{AIME'24} & \textbf{AQuA-R.} \\
    \hline
     R1-Distill & 0.98 & 0.98 & 0.98 \\\hline
     QwQ-32B & 0.98 & 0.98 & 0.98 \\\hline
    \end{tabular}
    \caption{The similarity threshold empirically chosen for Slim-SC (RP).
    }
    \label{tab:Slim-SC-RP}
\end{table}

\end{document}